\documentclass[sigconf]{acmart}

\setcopyright{acmcopyright}
\acmConference[FDG'19]{FDG}{August 26-30, 2019}{San Luis Obispo, CA, USA}


\begin{document}
\title{From Non-Paying to Premium: Predicting User Conversion in Video Games with Ensemble Learning}
\renewcommand\shorttitle{Predicting User Conversion in Video Games with Ensemble Learning}

\author{Anna Guitart}
\affiliation{%
\institution{Yokozuna Data, a Keywords Studio}
\postcode{102-0074} 
\city{Tokyo} 
\streetaddress{Chiyoda 5F, Aoba No. 1 Bldg.} 
\country{Japan}
}
\email{aguitart@yokozunadata.com}

\author{Shi Hui Tan}
\affiliation{%
\institution{Yokozuna Data, a Keywords Studio}
\postcode{102-0074} 
\city{Tokyo} 
\streetaddress{Chiyoda 5F, Aoba No. 1 Bldg.} 
\country{Japan}
}
\email{stan@yokozunadata.com}
\author{Ana Fern\'andez del R\'{i}o} 
\affiliation{%
\institution{Yokozuna Data, a Keywords Studio}
\postcode{102-0074} 
\city{Tokyo} 
\streetaddress{Chiyoda 5F, Aoba No. 1 Bldg.} 
\country{Japan}
}
\affiliation{%
\institution{Dpto. F\'isica Fundamental, UNED}
\postcode{102-0074} 
\city{Madrid} 
\streetaddress{Chiyoda 5F, Aoba No. 1 Bldg.} 
\country{Spain}
}
\email{afdelrio@yokozunadata.com}

\author{Pei Pei Chen}
\affiliation{%
\institution{Yokozuna Data, a Keywords Studio}
\postcode{102-0074} 
\city{Tokyo} 
\streetaddress{Chiyoda 5F, Aoba No. 1 Bldg.} 
\country{Japan}
}
\email{ppchen@yokozunadata.com}

\author{\'{A}frica Peri\'a\~{n}ez}
\affiliation{%
\institution{Yokozuna Data, a Keywords Studio}
\postcode{102-0074} 
\city{Tokyo} 
\streetaddress{Chiyoda 5F, Aoba No. 1 Bldg.} 
\country{Japan}
}
\email{aperianez@yokozunadata.com}

\renewcommand{\shortauthors}{Anna Guitart, Shi Hui Tan, Ana F. del R\'{i}o, Pei Pei Chen and \'{A}frica Peri\'a\~{n}ez}

\copyrightyear{2019}
\acmYear{2019}
\acmConference[FDG '19]{The Fourteenth International Conference on the Foundations of Digital Games}{August 26--30, 2019}{San Luis Obispo, CA, USA}
\acmBooktitle{The Fourteenth International Conference on the Foundations of Digital Games (FDG '19), August 26--30, 2019, San Luis Obispo, CA, USA}
\acmPrice{15.00}
\acmDOI{10.1145/3337722.3341855}
\acmISBN{978-1-4503-7217-6/19/08}

\begin{abstract}
Retaining premium players is key to the success of free-to-play games, but most of them do not start purchasing right after joining the game. By exploiting the exceptionally rich datasets recorded by modern video games---which provide information on the individual behavior of each and every player---survival analysis techniques can be used to predict what players are more likely to become paying (or even premium) users and when, both in terms of time and game level, the conversion will take place. 

Here we show that a traditional semi-parametric model (Cox regression), a random survival forest (RSF) technique and a method based on conditional inference survival ensembles all yield very promising results. However, the last approach has the advantage of being able to correct the inherent bias in RSF models by dividing the procedure into two steps: first selecting the best predictor to perform the splitting and then the best split point for that covariate.

The proposed conditional inference survival ensembles method could be readily used in operational environments for early identification of premium players and the parts of the game that may prompt them to become paying users. Such knowledge would allow developers to induce their conversion and, more generally, to better understand the needs of their players and provide them with a personalized experience, thereby increasing their engagement and paving the way to higher monetization. \end{abstract}


\maketitle

\section{Introduction}
\label{intro}
The recent paradigm change in video games---now games are always-online or have an online playing option---has driven a change in game monetization. A new business model has emerged: free-to-play or freemium games that can be acquired and played for free and only charge users for additional in-game content. Today a vast majority of mobile games follow this pricing strategy \cite{annie,Monetization}, and even traditional PC and platform games are relying more and more on extra contents to be purchased online as a source of revenue.

Identifying and retaining high-value players is crucial for successful monetization, especially in the case of freemium games \cite{perianez2016churn}. Previous research along these lines focused on predicting lifetime value (the amount a player will spend on purchases before leaving the game) \cite{Sifa2018CustomerLV,chen2018ltv} and churn---by trying to foresee what players are going to leave the game \cite{ding,kawale,hadiji,rothenbuehler2015hidden,runge2014} and when they are going to do it \cite{perianez2016churn,GameBigData,cig2018competition,chen2019competition}. The main idea behind these works is that pinpointing premium players who are likely to churn would allow developers to take steps to increase their lifetime in the game, since retention strategies are usually cheaper than acquisition campaigns~\cite{Monetization}.

In this paper we entertain a similar idea: that the ability to predict what players have the potential to become paying users (PUs) and when (or at what game level) they are more likely to start purchasing would allow developers to take steps to induce their conversion. And this ability could lead to a significant increase in monetization, since getting users to purchase remains challenging even for big games: up to 70\% of players quit the game without having spent any money \cite{forbesConversions}. For example, a game may be very engaging (very high retention rates) but present poor user conversion rates. 

Once the game already has a base of users actively engaged in purchasing and/or continuous conversions from non-PUs to PUs player retention strategies come into play. 

Another related issue is spotting the existing PUs who have the potential to become \emph{whales} (top spenders). These are the most valuable players, typically providing up to 50\% of the total revenue of the game despite accounting for less than 1\% of the total number of players \cite{whales}, 
and thus their early identification is of the utmost importance. 

To tackle this conversion prediction problem, we will apply survival analysis, a set of statistical methods used to estimate the time it takes for a certain event of interest---in our case, becoming a PU---to happen. 
We will explore three different approaches (the traditional Cox regression model, a random survival forest (RSF) technique and a method based on conditional inference survival ensembles) and provide predictions in terms of the number of days, in-game levels and cumulative playtime before a certain user becomes a PU. It is worth noting that, contrary to churn prediction in casual games (where the churn definition is not straightforward \cite{hadiji,perianez2016churn}) in this case the event of interest is clearly defined: it occurs the moment the player makes a purchase.

The prediction of conversion times has been thoroughly investigated in other fields, such as e-commerce \cite{cui2018modelling} or medicine and healthcare \cite{wu2018discrimination}, with some works also making use of survival analysis techniques. For instance, in \cite{Ji2017} a conversion prediction model, together with a recommendation system, is proposed in connection to e-commerce websites, while the authors of \cite{wang2013time} modeled career switches using the proportional hazards model.

In the context of video-games, previous research about conversion treats it as a binary classification problem \cite{sifa2015predicting}, where players are divided into potential and non-potential PUs through traditional machine learning techniques, such as support vector machines, decision trees and random forests. 

\subsection{Our Contribution}
Previous studies have already shown the application of survival analysis to video games for predicting churn \cite{perianez2016churn,GameBigData} but, to the best of our knowledge, this is the first paper using a survival approach to predict conversion times in the context of video games.

\section{Survival Analysis Models}
\label{models}
Survival analysis \cite{clark} was introduced to address time-to-event regression problems characterized by having incomplete or partially labeled data. This set of methods focus on estimating the remaining lifetime of an individual until a specific event happens, given a set of predictor (explanatory) variables. Traditionally, the event of interest used to be death or organ failure, as these techniques were first applied in the biological and medical fields \cite{hougaard1999fundamentals}. In this work, the event of interest is \emph{becoming a PU}. The time to the event of interest cannot be determined until it happens and hence not all individuals can be labeled, a situation known as \emph{censoring}. 
A special type of time-to-event models considers the existence of \emph{competing risks} \cite{Prentice}, events which impede the observation or affect the probability of occurrence of the event of interest.

The outcome of survival models is the survival probability curve for each individual, which indicates the probability that the event has \emph{not} happened yet (i.e. that the user is still \emph{alive}) at a certain time point.

However, for a more intuitive understanding, in this study we will depict the cumulative incidence function, which gives the probability that the event of interest---becoming a PU---\emph{does happen}.

The predicted time-to-event is derived from the survival curves: it is identified with the \emph{median survival time}, the time for which survival probability gets down to 50\%.
The survival function $S(t)$ is related to the hazard function $h(t)$, defined as the ratio of the probability density function $P(t)$ to the survival function: 
\begin{equation}
h(t) = \frac{P(t)}{S(t)}.
\label{hazard}
\end{equation}

In this paper we focus on comparing the performance of a semi-parametric model (the Cox proportional hazards model) to that of more recent survival ensemble techniques, such as the \emph{conditional inference survival ensembles} and \emph{random survival forest} methods. For the latter, we also tested the inclusion of competing risks. These models are presented in the following sections.

\subsection{Cox Regression}
\label{CoxReg}
The Cox proportional hazards or Cox regression model~\cite{Cox1972,cox1984analysis,david1972regression} is a survival model that assumes a multiplicative relation between covariates and hazard:
\begin{equation}
S^{\rm Cox}(t|x_i) = \exp\bigl(-h_0(t)\exp(\beta^T x_i) \bigr).
\label{CoxFormula}
\end{equation}
Here $h_0$ is the baseline hazard function, $\beta$ is an unknown vector of regression coefficients (parameters) and $x_i$ are the covariates for each individual $i$, with $i=1,\ldots,M$.

Cox regression is a very popular method and is frequently used in survival analysis due to its flexibility as a semiparametric model. The hazard function is estimated in a distribution-free manner from the data, and there exists a linear-exponential parametric relationship between the predictors and the outcome.

\subsection{Conditional Inference Survival Ensembles}
\label{CondInfSurvEns}

The conditional inference survival ensembles (also known as \emph{conditional inference forest}) model is a fully non-parametric tree-based method used in survival analysis. It is based on the Breiman random forest \cite{breiman2001random}, but uses conditional inference trees (instead of the usual decision trees) as base learners \cite{Hothorn06unbiasedrecursive}. The splitting at each node is performed in two steps: (1) the optimal split variable is selected based on its correlation with the output, and (2) the best split point for that covariate---the one that maximizes the survival difference among daughter nodes---is determined using two-sample linear statistics.

Conditional inference forests use a weighted Kaplan--Meier estimate \cite{hothorn2004bagging, mogensen2012evaluating} to construct the survival function~\cite{mogensen2012evaluating,perianez2016churn}:
\begin{equation}
S^{\rm conditional}(t|x_i) = \prod \Biggl(1 - \frac{\sum_{n=1}^{N} T_n(dt,x_i)}{\sum_{n=1}^{N} Q_n(t,x_i)} \Biggr),
\label{condEnsFormula}
\end{equation}
where $n = 1, \ldots, N$, with $N$ the number of trees within the ensembles, and $x_i$ are the covariates
for the $i$th subject, with $i=1,\ldots,M$.
In the node where $x_i$ is located, $T_n$ represents the uncensored events until time $t$, and $Q_n$ stands for the number of individuals at risk at~$t$.\looseness=-1 

\subsection{Random Survival Forest}
\label{SRF}

The random forest algorithm was first described in \cite{breiman2001random}. It consists of an ensemble of decision trees trained using bootstrap samples from the total set, with selection of the splitting variable at each node being random. The split point is taken as the one that maximizes a predefined splitting criteria (often, the Gini impurity measure \cite{breiman1984classification}). The selection of the split variable and split point is performed at the same step, which gives rise to a relatively biased model that favors variables with many possible split points.
The survival extension of this method is called \emph{random survival forest}~\cite{ishwaran2008random}. 

The ensemble is constructed using tree-based Nelson--Aalen estimators \cite{ishwaran2008random}:
\begin{equation}
H_n(t,x_i) = \int_{0}^{t}  \frac{T_n(dt,x_i)}{Q_n(t,x_i)}
\label{SRFestimators}
\end{equation}
and the ensemble survival function is
\begin{equation}
S^{\rm SRF}(t|x_i) = \exp\biggl(- \frac{1}{N}\sum_{n=1}^{N} H_n(t,x_i)\biggr),
\label{SRFFormula}
\end{equation}
where the variables have the same meaning as in \eqref{condEnsFormula}.

This model, as the previously described ensemble model, is fully non-parametric, which offers an advantage over other approaches.

\subsection{Random Survival Forest with Competing Risks}
\label{SRF_cr}

\begin{figure}[ht!]
  \centering
  \includegraphics[width=\columnwidth]{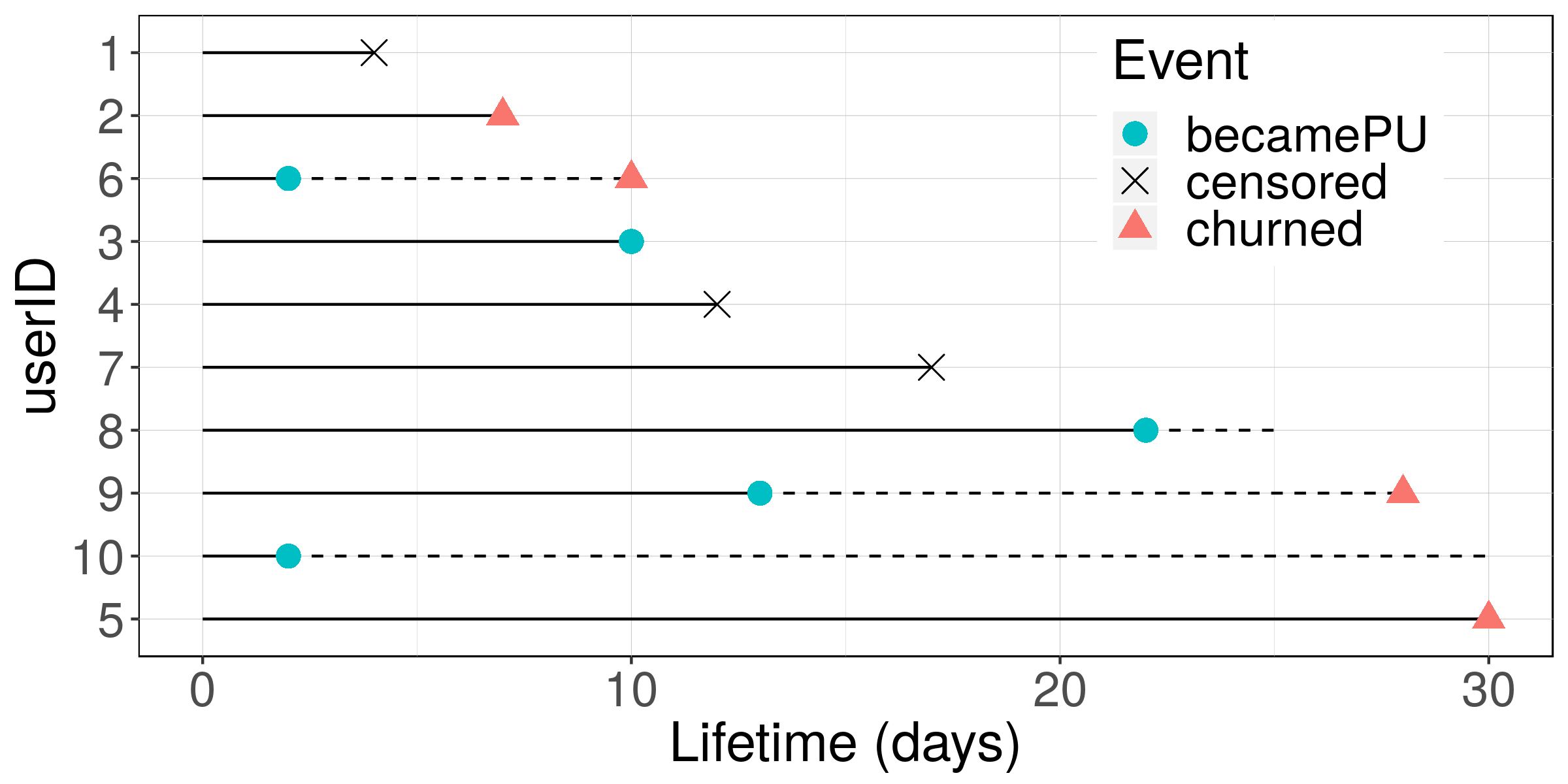} 
\caption{
Example of right-censored data (for 10 users over 30 days of lifetime) considering churn as competing risk. Players may become PUs (circles) or churn (triangles) at some point. If neither of these two events occur within the observation period, then the data is censored (crosses).}
\label{censoring}
\end{figure}

This is an extension of the random survival forest method explained in the previous section in which competing risks are considered~\cite{ishwaran2014random}.
Throughout this work, we assume the main reason that prevents the event of interest from happening (i.e.\ that prevents players from becoming PUs) is a lack of interest in purchasing. However, now we will also take into account the fact that players may not become PUs because they \emph{churn} (leave the game) before. Thus, we have two events of interest that conflict with each other: becoming a PU and churning, see Figure \ref{censoring}. We will only consider player information until one of these two events occur, as once a user has churned she obviously cannot become a PU anymore.

Including competing risks affects the splitting rules used to grow the survival trees, and the values computed in each terminal node of the ensemble become event-specific~\cite{ishwaran2014random}.

For random forests with competing risks, a competing risk tree is grown for each bootstrap sample and the node is split using the best covariate---the one that maximizes the competing risk splitting rule.

The cumulative event-specific hazard function for each event $j$ considering a Nelson--Aalen estimator is given by

\begin{equation}
H_{nj}(t,x_i) = \int_{0}^{t}\frac{T_{nj}(dt,x_i)}{Q_n(t,x_i)} = \sum_{k=1}^{m(t)} \frac{d_{nj}(t_k,x_i)}{Q_n(t_k,x_i)},
\label{SRFcrEstimators}
\end{equation}
where $m(t) = \max\{k: t_k \leq t\}$ and {\thickmuskip=3mu minus 1mu$d_{nj}(t_k)=\sum_{i=1}^{M}I(T_i=t_k,\delta_i=j)$} is the number of type-$j$ events at time $t_k$ for all individuals $i$, with $I$ being the corresponding event indicator. (The total number of events occurring at time $t_k$ is denoted as $d_{n}=\sum_{j}\delta_{jn}(t_k)$.)

\section{Datasets}
\label{dataset}
The work presented in this article focuses on the analysis of two datasets from two different game titles: Age of Ishtaria (hereafter, AoI) and Grand Sphere (hereafter, GS). Both titles are role-playing card battle games very popular in Japan and developed by Silicon Studio, with the first one having a larger number of active players (although they are very similar). 
Data comprises daily records of the daily activity of each player (playtime, actions, sessions, etc.) and was collected between January 2015 and February 2017 for AoI and between June 2017 and May 2018 for GS.  
During these periods, neither of the games experienced major changes that might have influenced the data, see~\cite{cig2018competition,chen2019competition}.

\begin{figure*}[ht!]
  \centering
  \includegraphics[width=0.3\textwidth]{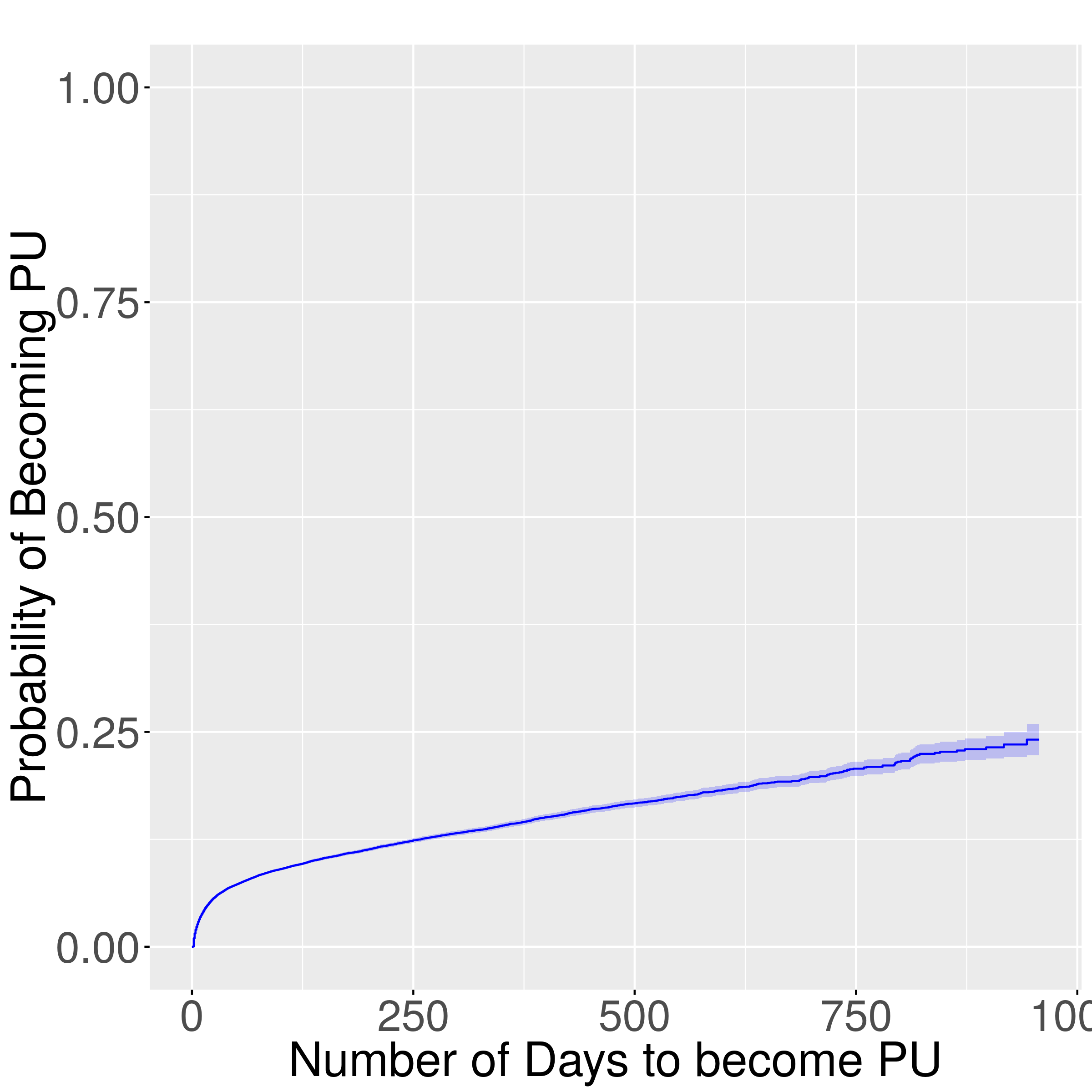} 
  \includegraphics[width=0.3\textwidth]{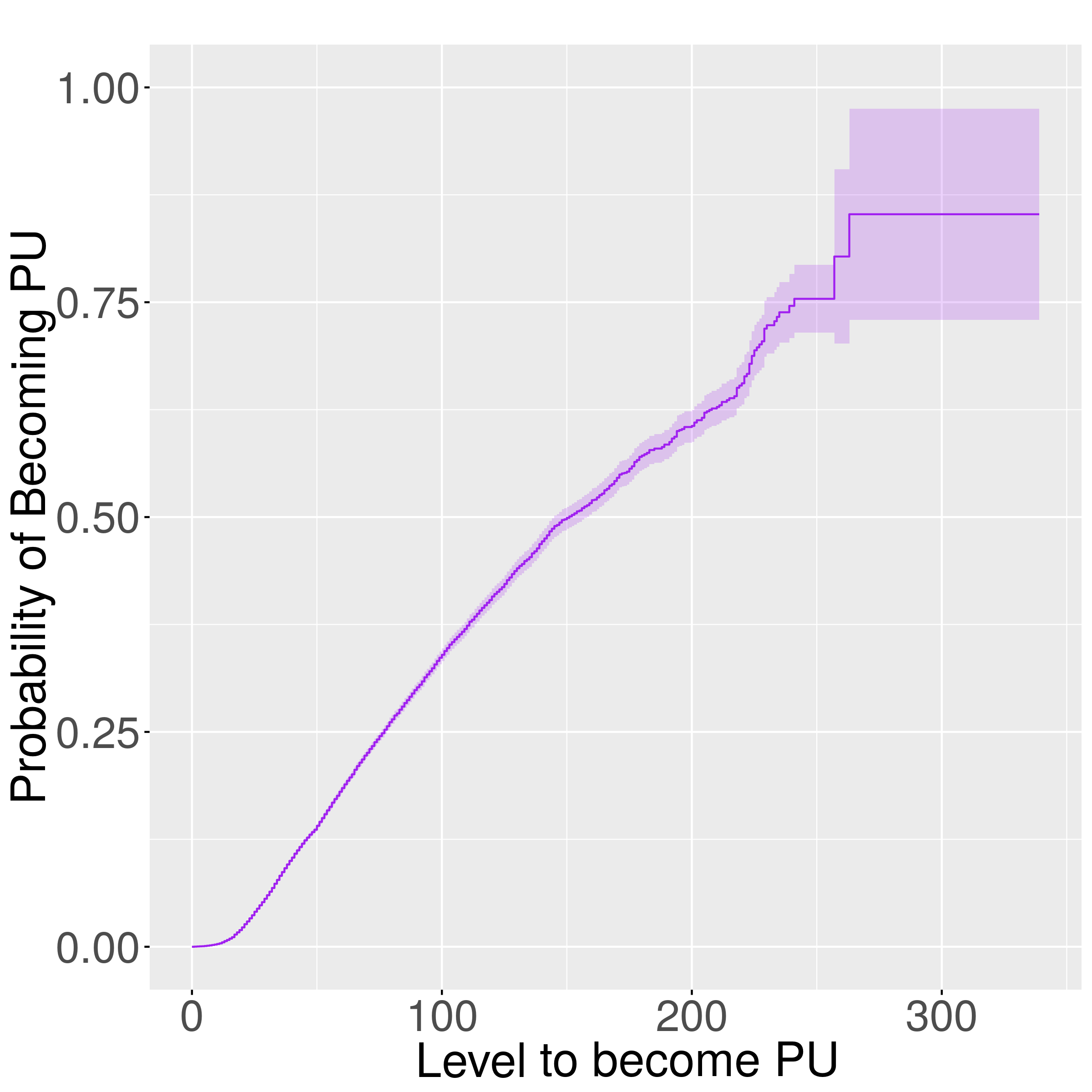} 
  \includegraphics[width=0.3\textwidth]{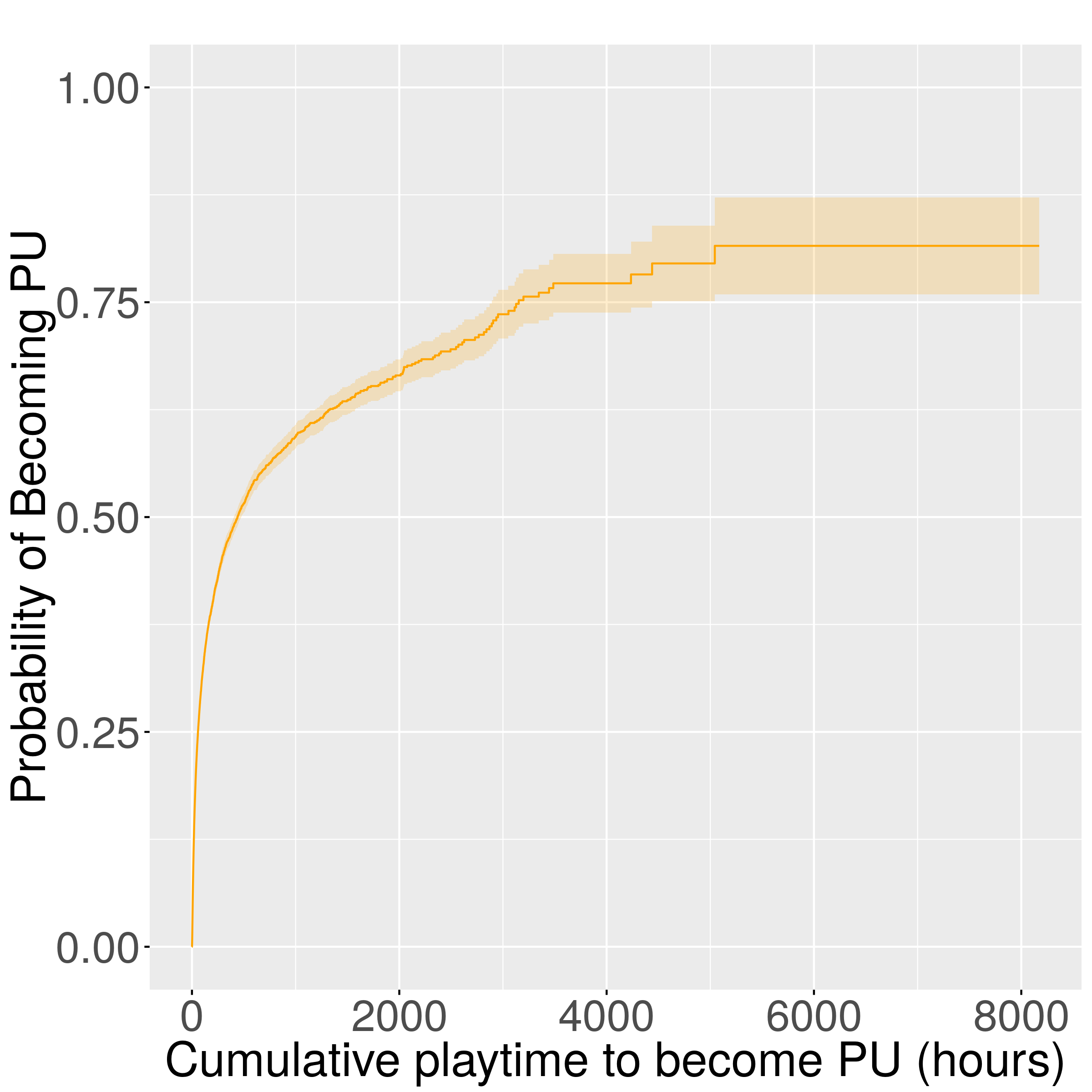}\\
  \includegraphics[width=0.3\textwidth]{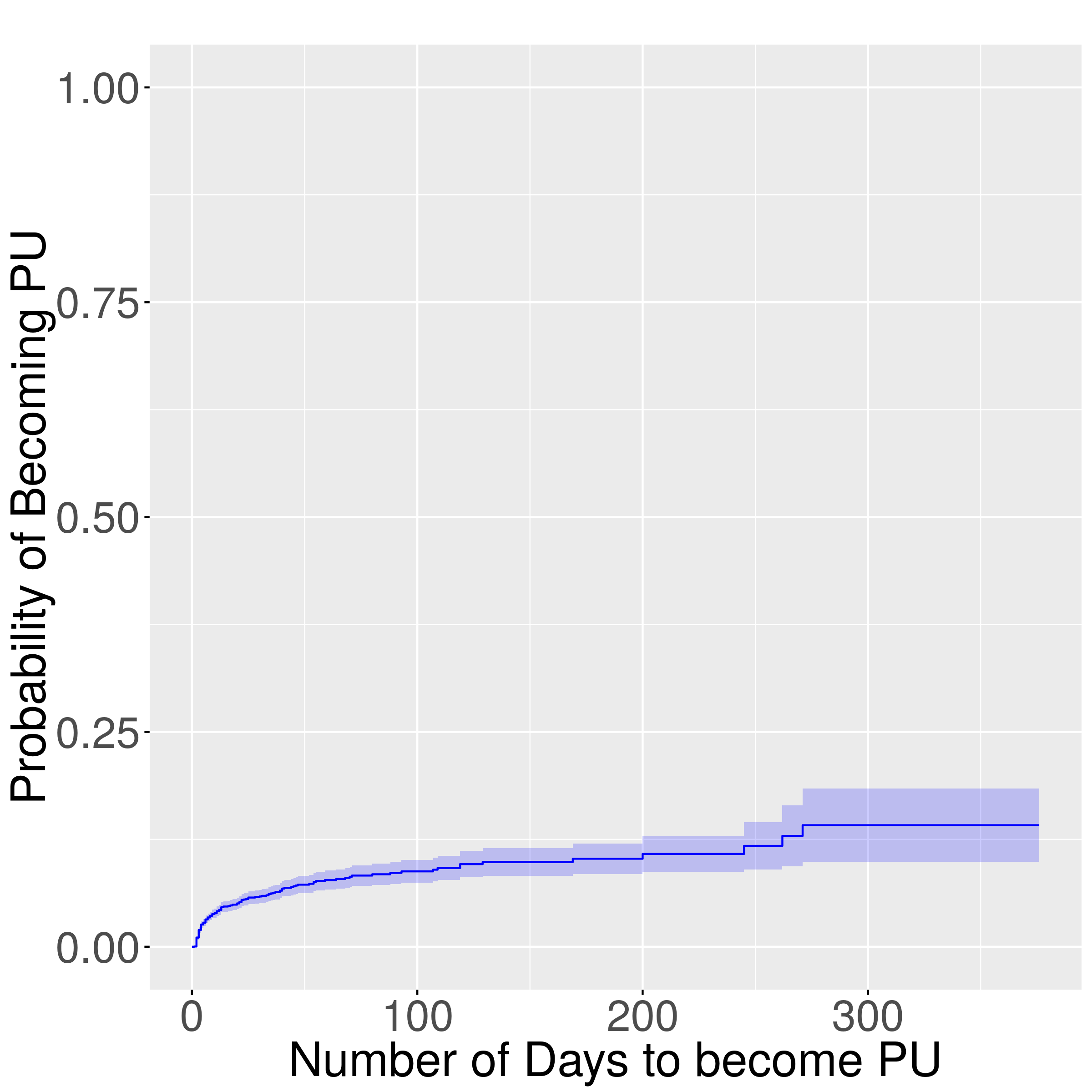} 
  \includegraphics[width=0.3\textwidth]{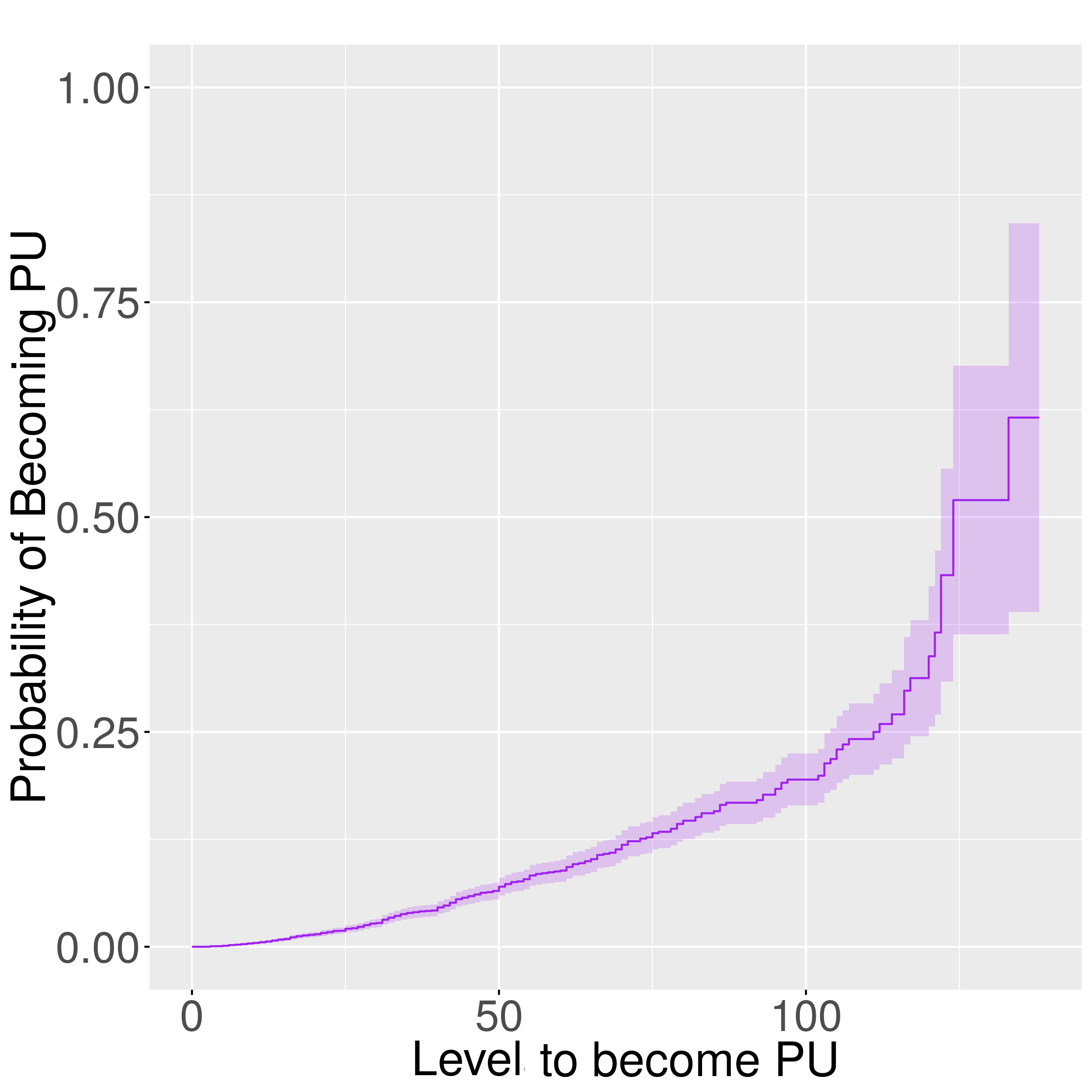} 
  \includegraphics[width=0.3\textwidth]{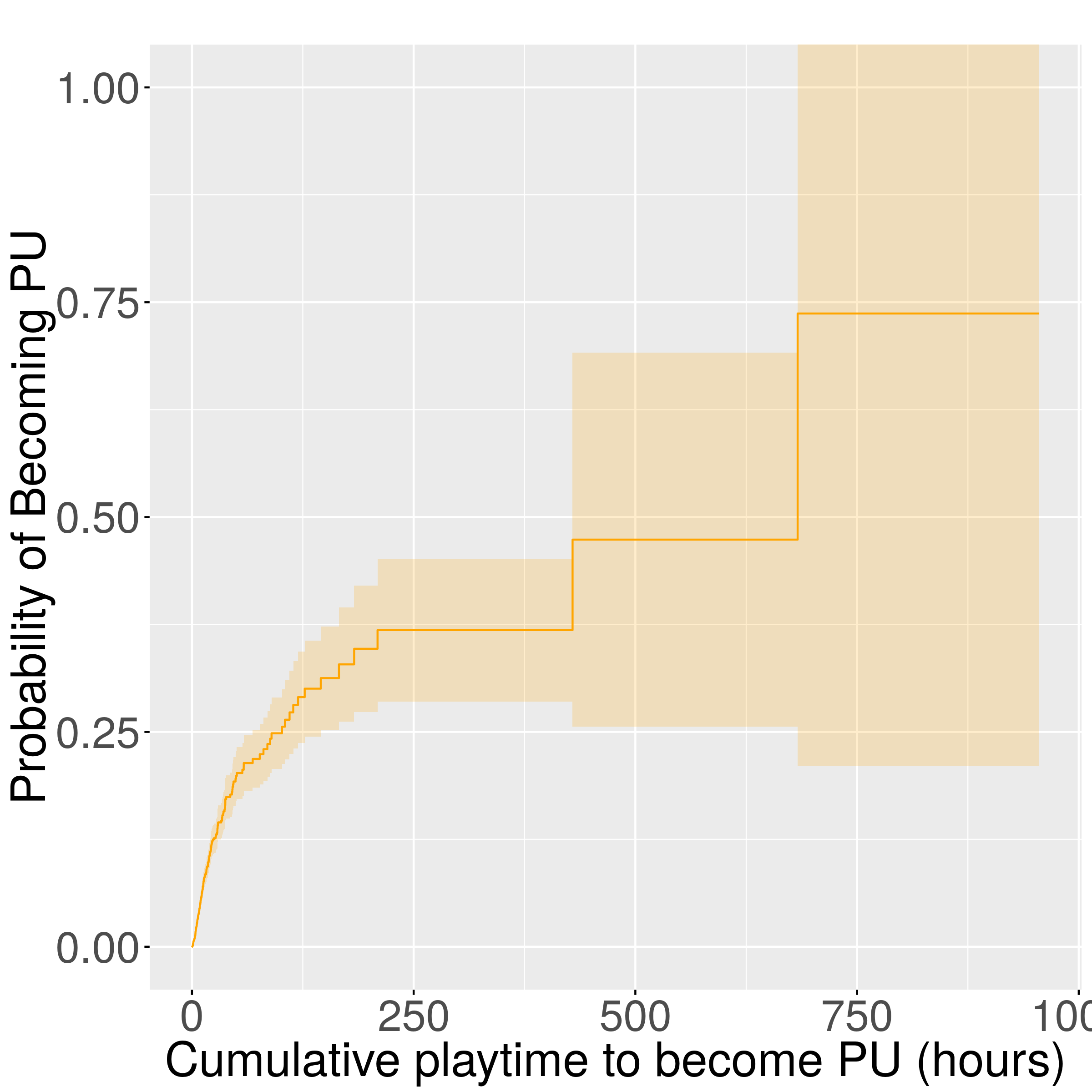}\\
\caption{Cumulative incidence functions, showing the probability of becoming a PU as a function of the number of days since registration (left), game level (center) and cumulative playtime (right) for all players in the games AoI (top) and GS (bottom). The shaded area represents the 95\% confidence interval.}
\label{KME}
\end{figure*}

\begin{figure*}[ht!]
  \centering
  \includegraphics[width=0.3\textwidth]{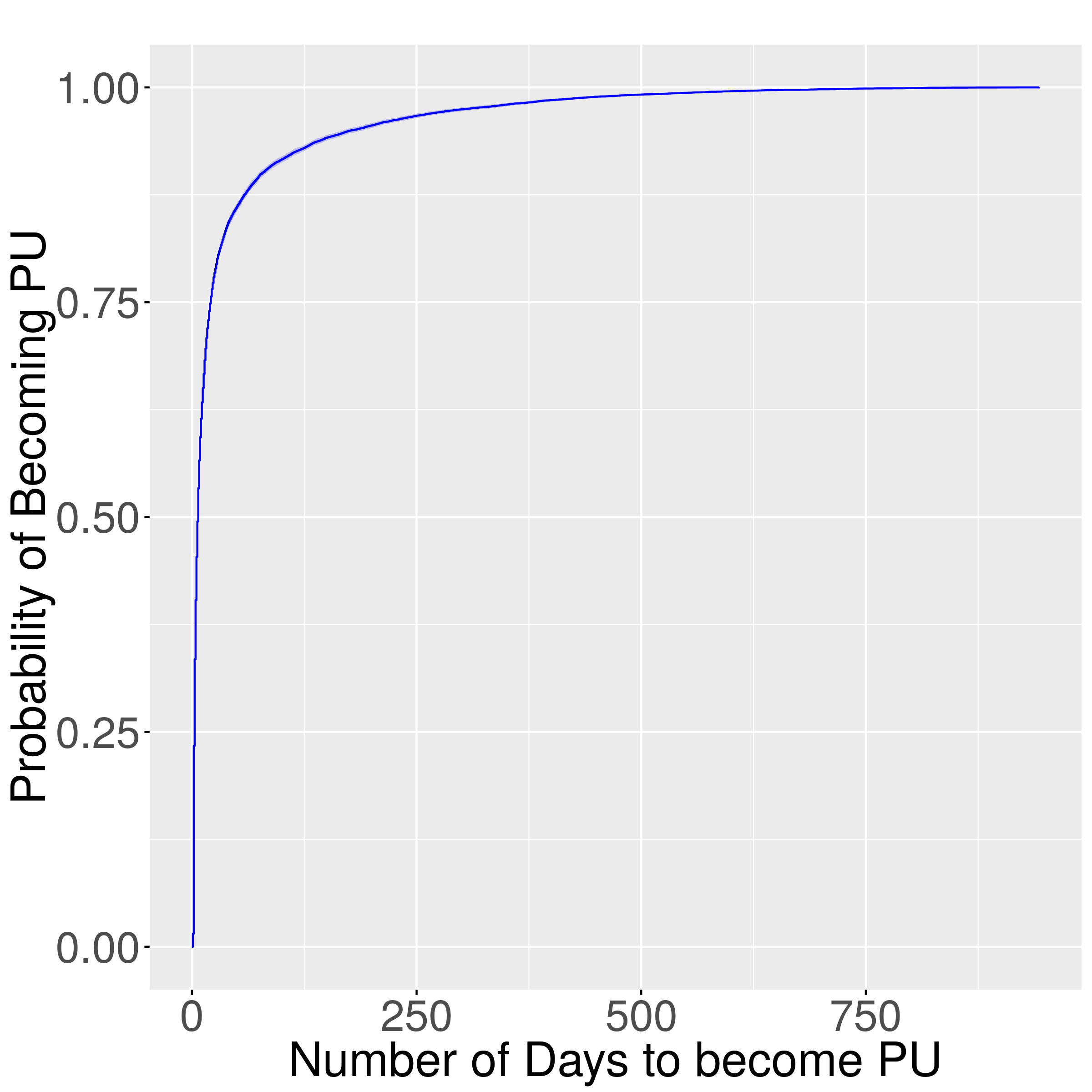} 
  \includegraphics[width=0.3\textwidth]{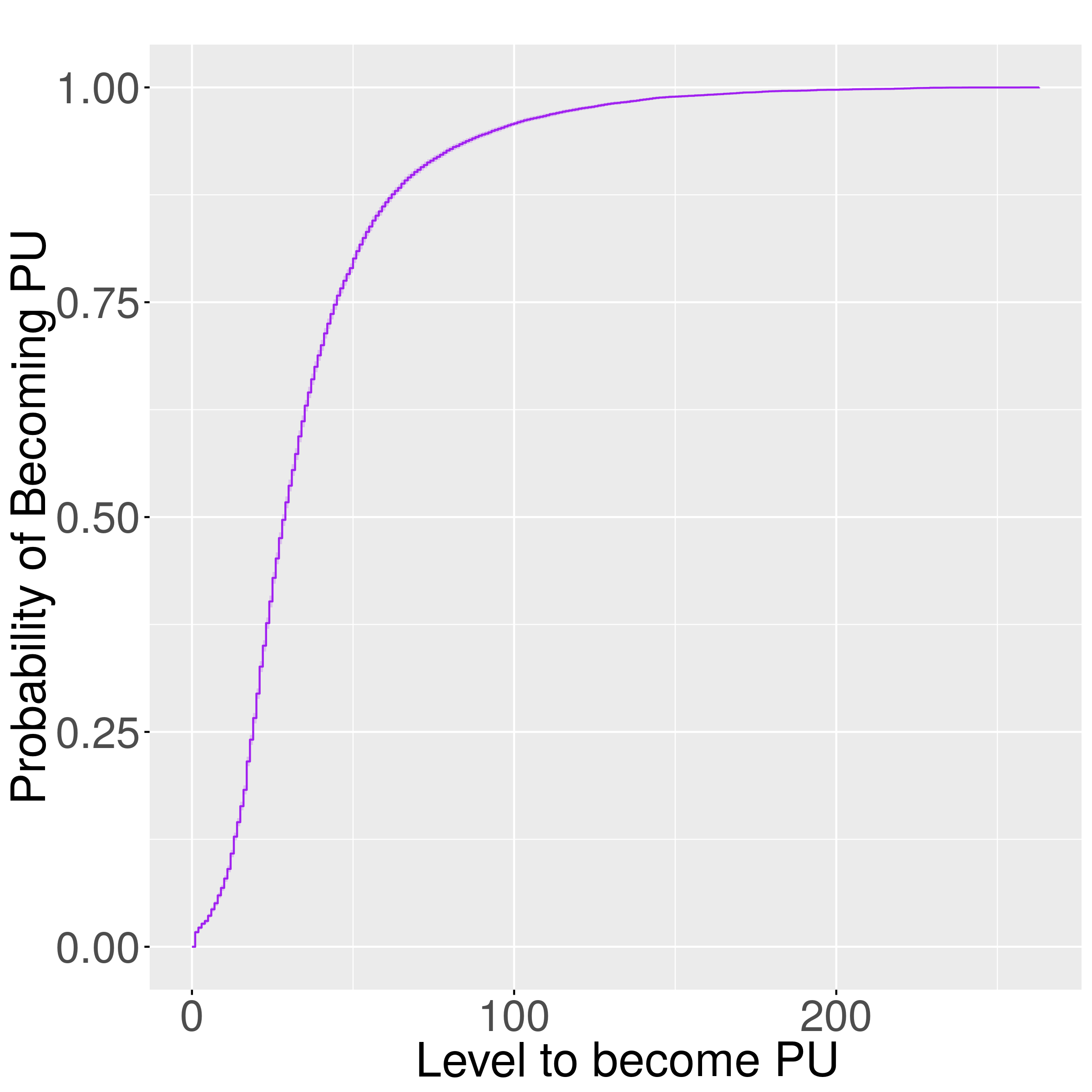} 
  \includegraphics[width=0.3\textwidth]{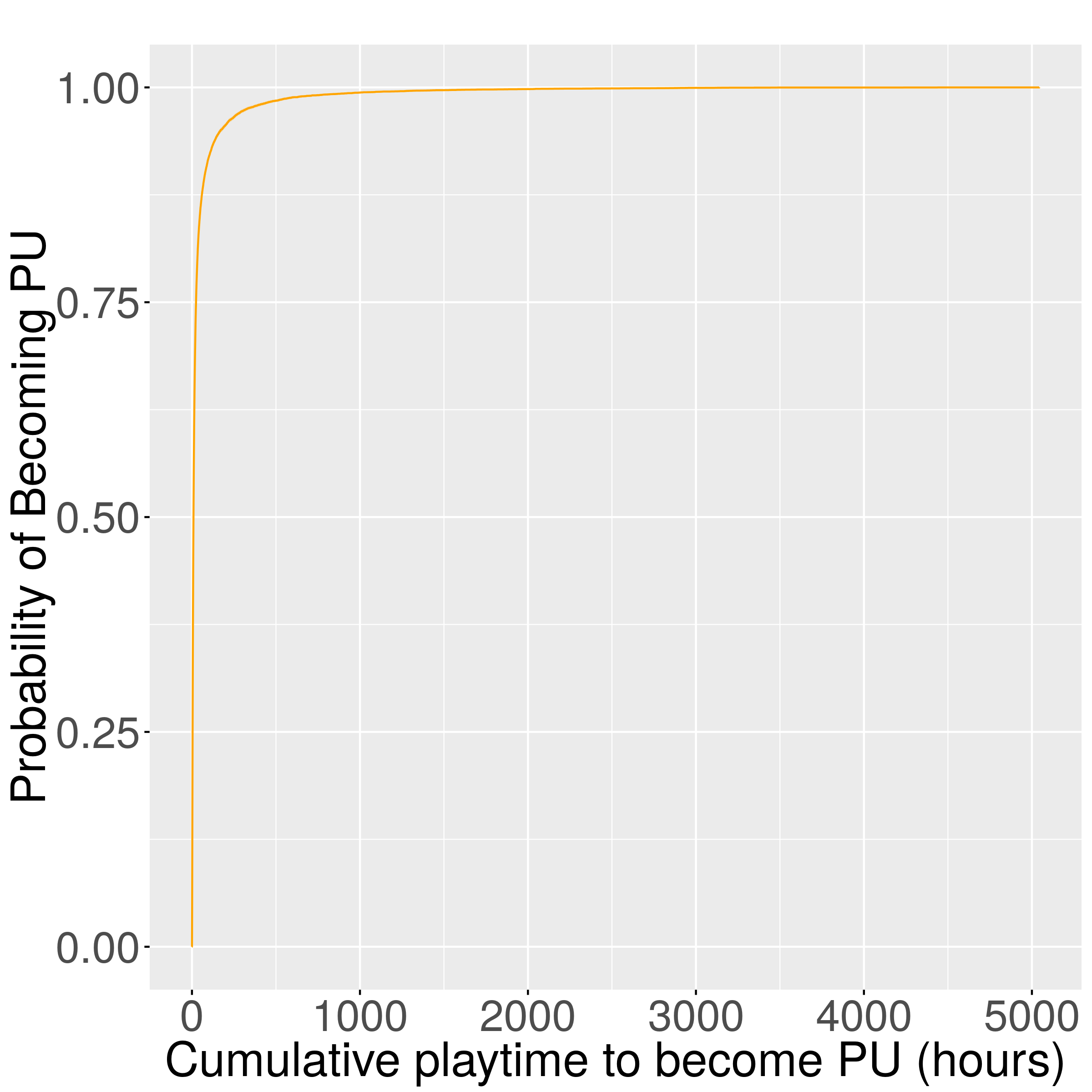}\\
  \includegraphics[width=0.3\textwidth]{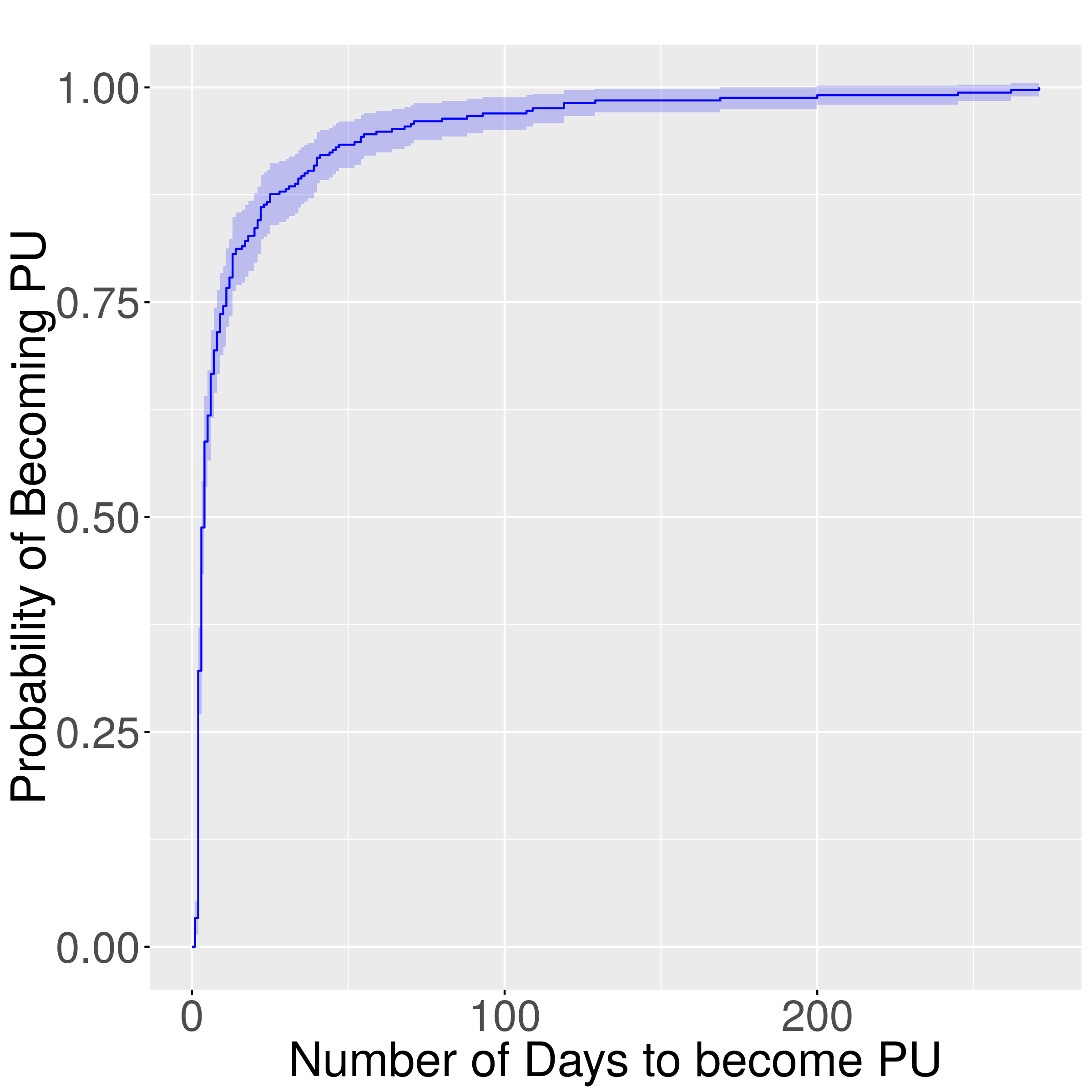} 
  \includegraphics[width=0.3\textwidth]{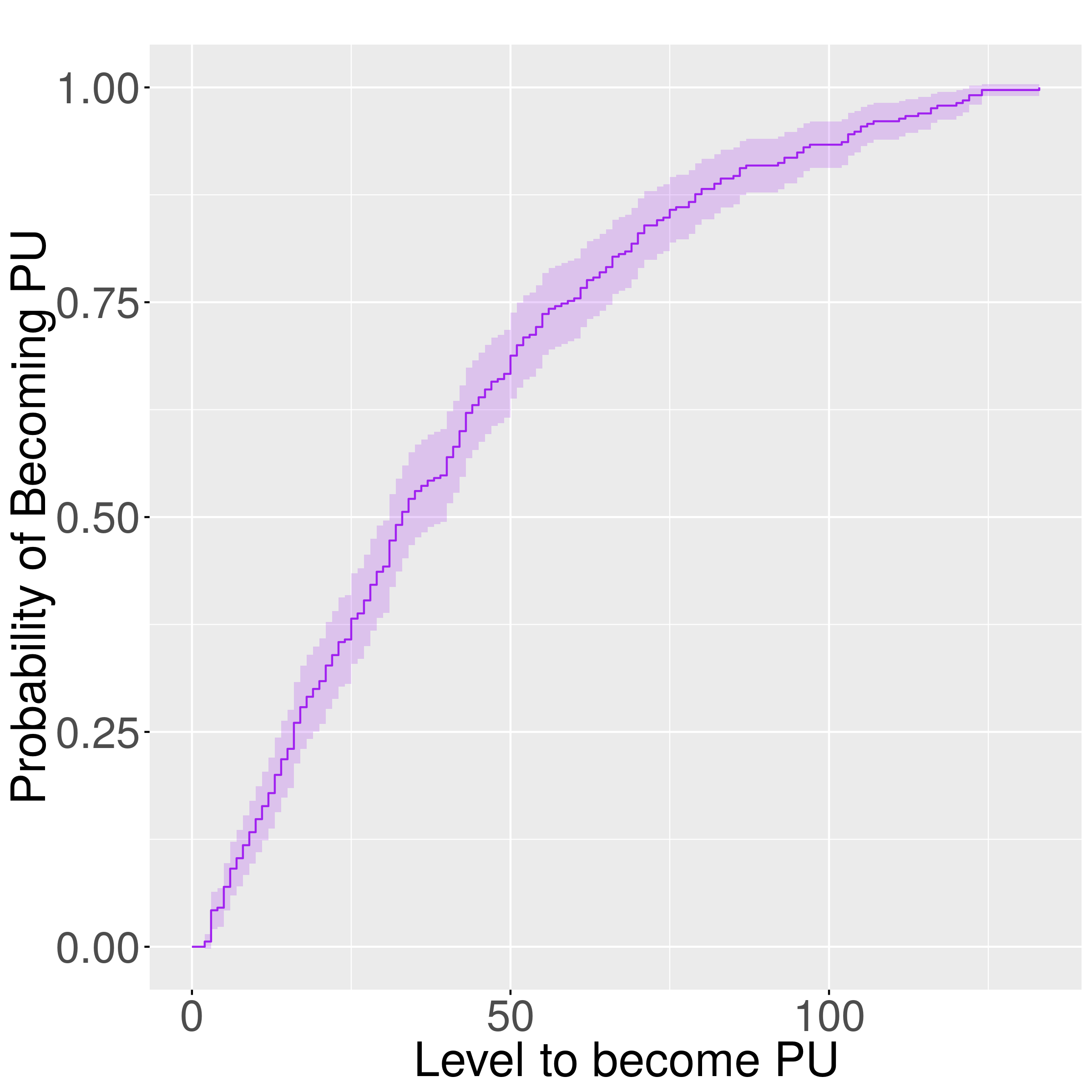} 
  \includegraphics[width=0.3\textwidth]{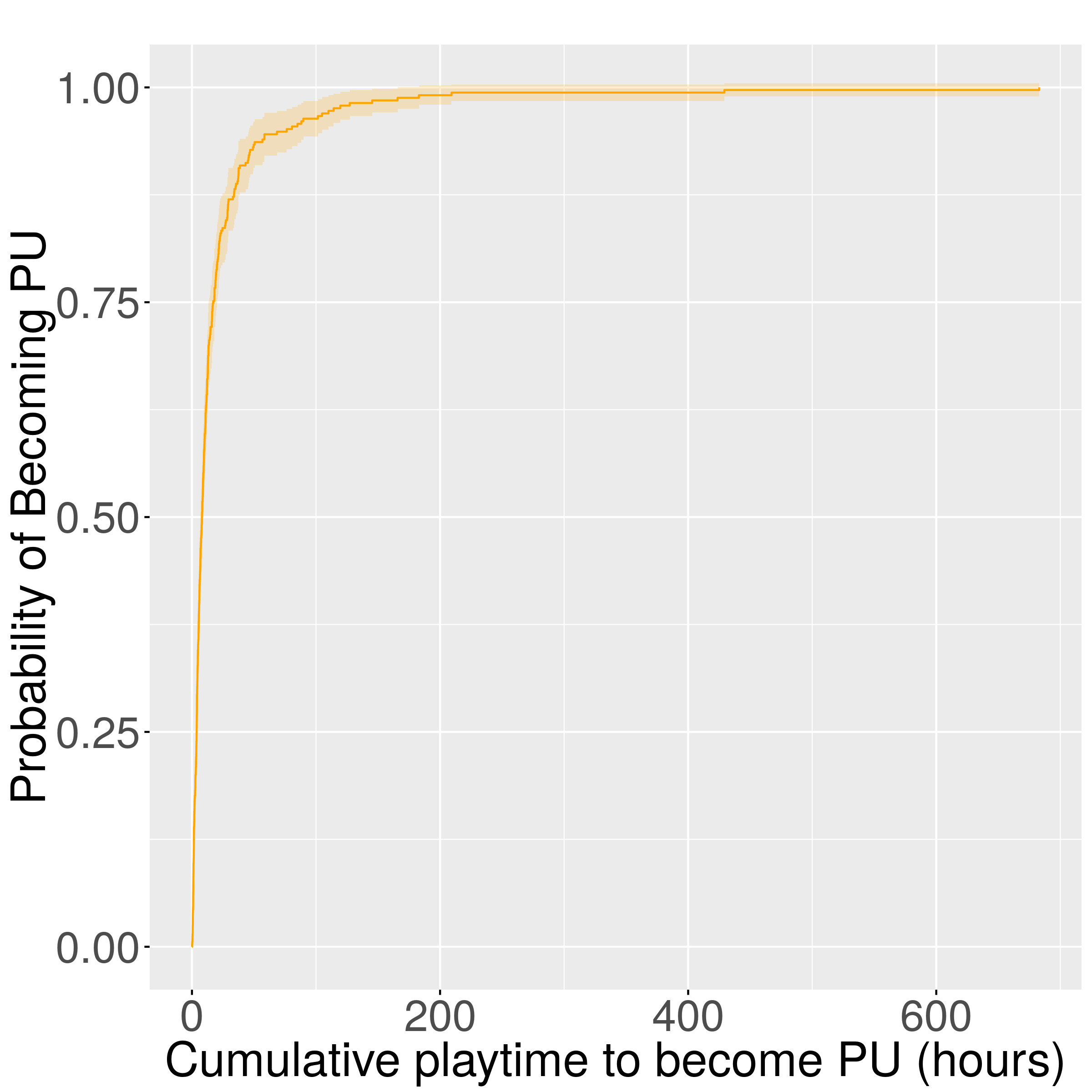}\\
\caption{Cumulative incidence functions, showing the probability of becoming a PU as a function of the number of days since registration (left), game level (center) and cumulative playtime (right), for PUs only, in the games AoI (top) and GS (bottom). The shaded area represents the 95\% confidence interval.}
\label{KME_onlyPU}
\end{figure*}

Only a small percentage of users will eventually become PUs, a pattern that can be observed in Figures~\ref{KME} and \ref{KME_onlyPU}. These figures show the \emph{inverse} of the Kaplan--Meier estimates for the probability of {\it surviving} as a non-paying user, i.e., they show the probability of becoming a PU in terms of the number of days, level achieved and accumulated playtime, both for the total population of players (Figure~\ref{KME}) and considering only PUs (Figure~\ref{KME_onlyPU}). Looking at the probability in terms of the number of days (Figure~\ref{KME}, left), we see that only around 25\% or less of all players end up becoming PUs. In the plots for the number of game levels (center) and cumulative playtime (right) to become a PU, final percentages are higher, as the few players who reach higher levels or longer playtimes are mostly PUs. This does not happen for the probability in terms of the number of days though: even if players stay in the game for a very long time, only a few of them will become premium users.

We considered only players who logged in at least 2 days in the game, thus discarding new players. In freemium games, every day there are typically many new registered users, most of whom will not connect a second day---they are \emph{one-time comers}. However, in operational settings, complete data from the first connection day is not available until the day has ended. Therefore, predicting the behavior of newcomers requires a different approach that is beyond the scope of this paper. By removing these new players, class imbalance is also reduced, as the vast majority of them will never become PUs. For non-newcomers, the percentage of PUs in our datasets was 5.32\% for AoI and  5.30\% for GS. 

Our sample comprised 30,000 users for AoI and 10,000 users for GS.

To perform the data splitting into train and test sets, we took random samples, ensuring that the proportion of PUs was similar in both sets; 30\% of players were assigned to the training set and the remaining 70\% constituted the test sample.

One of the aims of this exercise was to test if our models could provide accurate prediction results in an operational environment---where datasets can be huge---when trained with just a small subset of the total data. This is why we used a training set much smaller than the test set.

\subsection{Response Variables}
The implemented models were trained to predict the number of days to become a PU, the level at which each player will become a PU and the number of hours she will play until then. Similarly as in \cite{GameBigData}, we used the following predictor variables:
\begin{itemize}
    \item \textbf{Lifetime}: Number of days since the user's registration date.
    \item \textbf{Level}: Latest game level reached by the player.
    \item \textbf{Playtime}: Number of hours played by the user.
\end{itemize}
In all cases, the censored variable was whether the player became a PU or not. When including competing risks, there is an additional event to consider: whether the user churned before becoming a PU. For conversions, the event definition is straightforward: the event takes place as soon as the player makes her first purchase. In the case of churn, the definition is not as clear, and the event is usually assumed to happen after a certain inactivity period that may vary from game to game. This has been already discussed in depth in \cite{perianez2016churn,GameBigData,chen2018ltv}.

\subsection{Feature Selection}
We considered features not related to the peculiarities of the games and that can be measured in practically any title, as having game-independent features makes it easier to apply our research to real business environments. 
They were mainly based on playtime and actions/sessions, and several statistical operations (averaging playtime, etc.) were performed to obtain the final static features.
We also explored features related to user level, as most games have some measure of in-game progression (e.g.\ game or player level).
For each outcome---number of days, level, cumulative playtime---we selected the features that best modeled every output through a feature engineering process.

\section{Modeling}
\label{modeling}

\subsection{Model Specification}

For the ensemble methods (the conditional inference survival ensembles model and the random survival forest model, either with or without competing risks) we selected 900 trees to be used as base learners. 

As validation metrics, we used the root mean square logarithmic error (RMSLE) between the observed and predicted values, false positive rate (percentage of players in the validation sample who were predicted to become PUs but churned before doing so) 

and false negative rate (players who became PUs despite not being predicted to do so). Scatter plots of predicted vs.\ observed variables are also examined. 

\subsection{Results}
\label{results}

\begin{figure*}[ht!]
  \centering
  \includegraphics[width=0.30\textwidth]{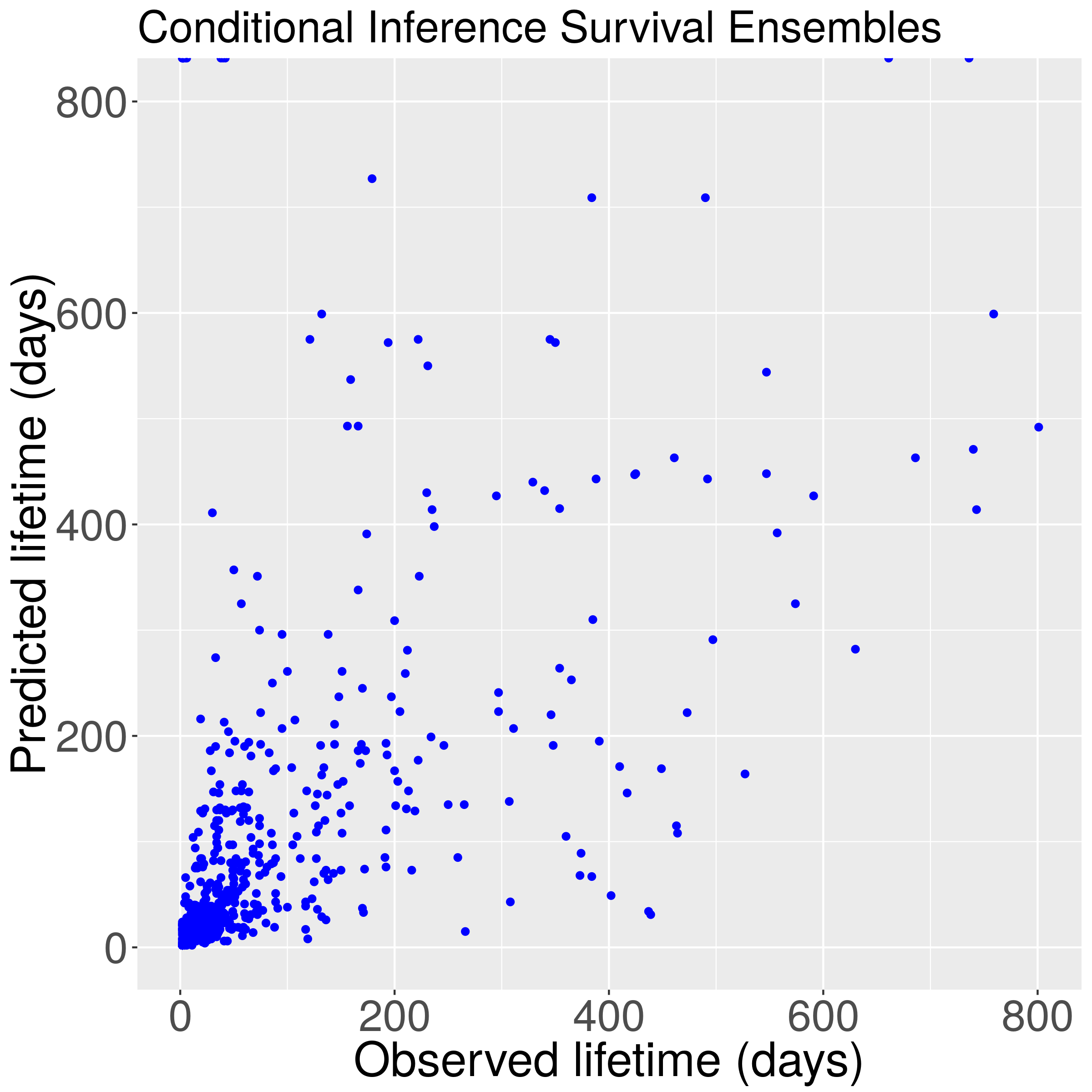} 
  \includegraphics[width=0.30\textwidth]{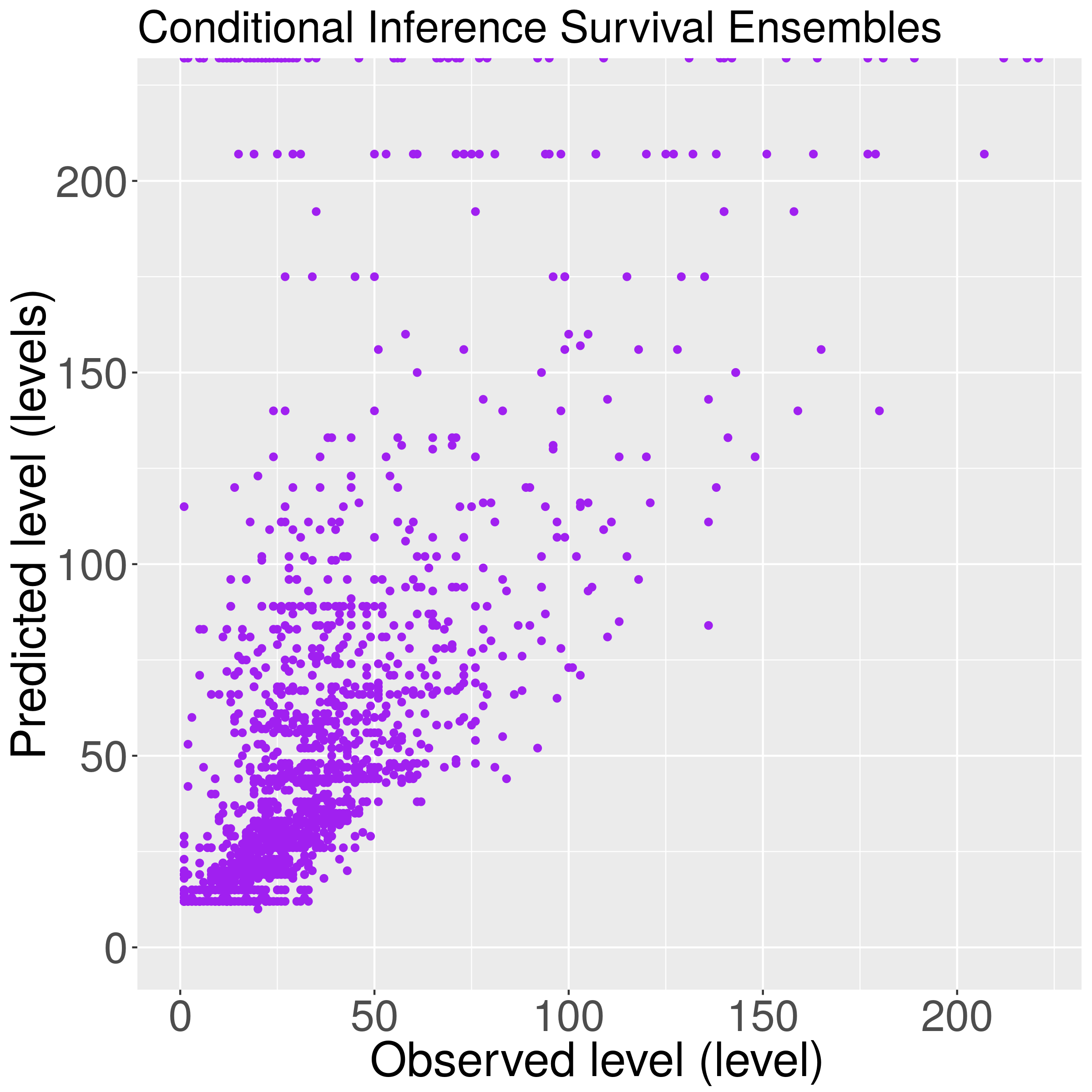} 
  \includegraphics[width=0.30\textwidth]{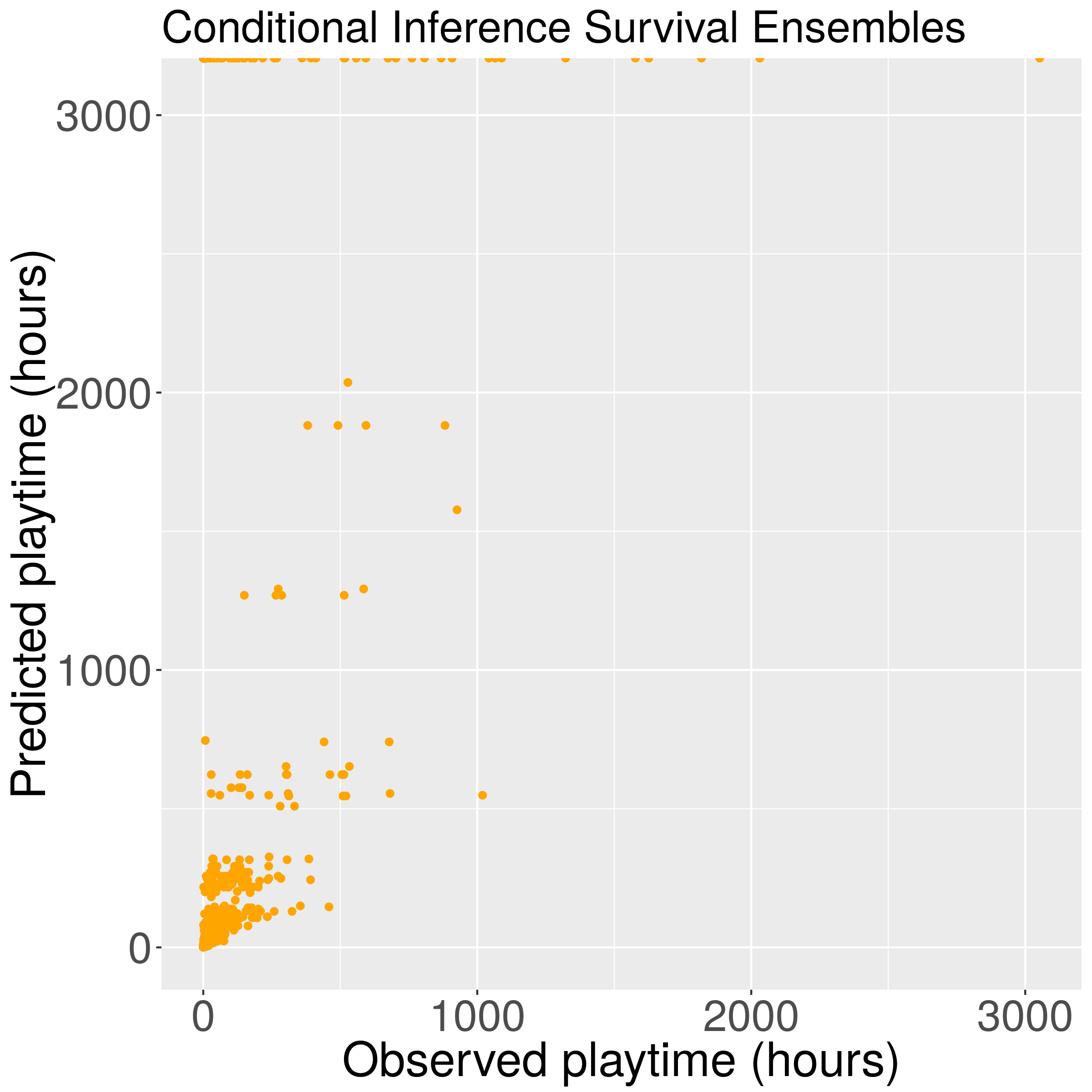}\\
  \includegraphics[width=0.30\textwidth]{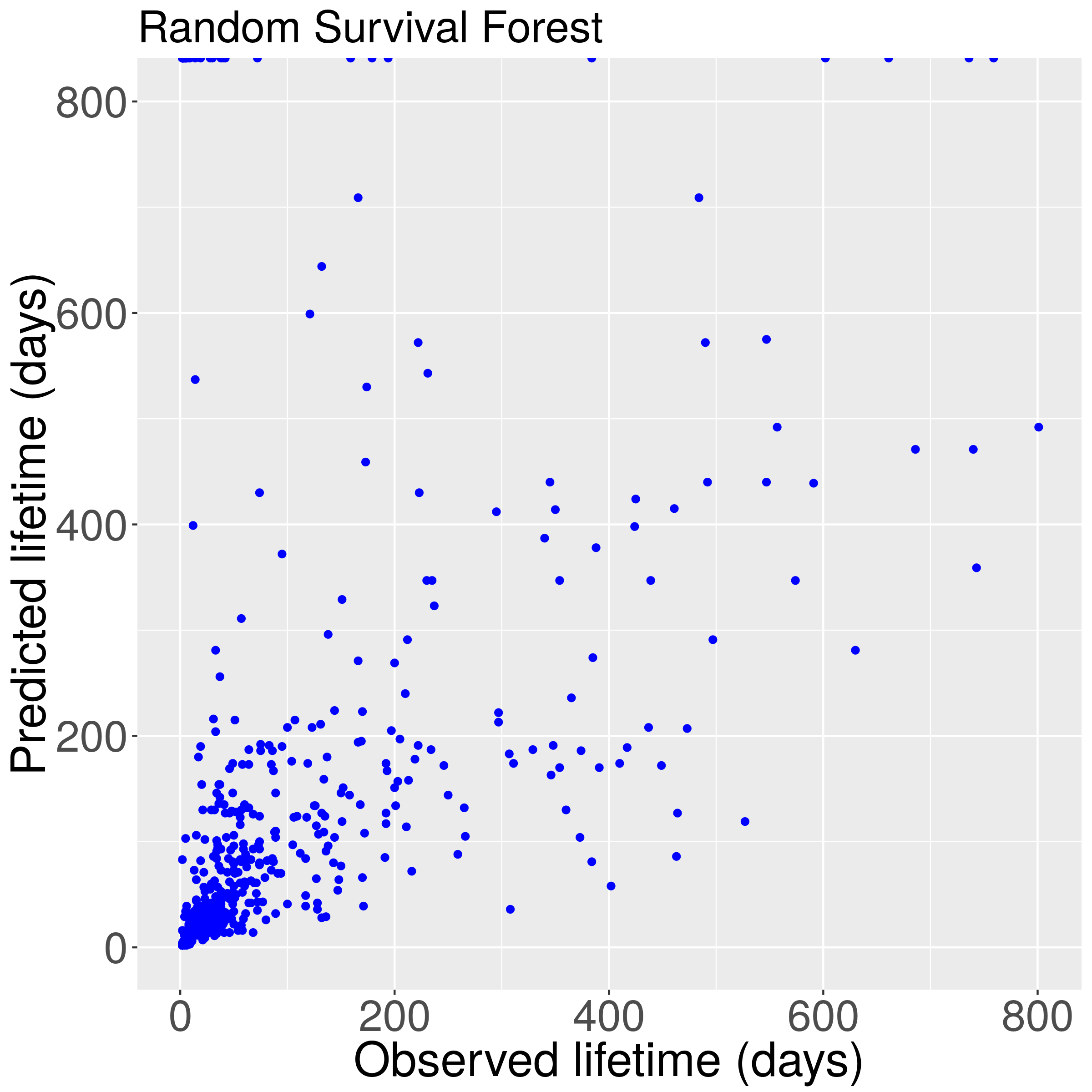} 
  \includegraphics[width=0.30\textwidth]{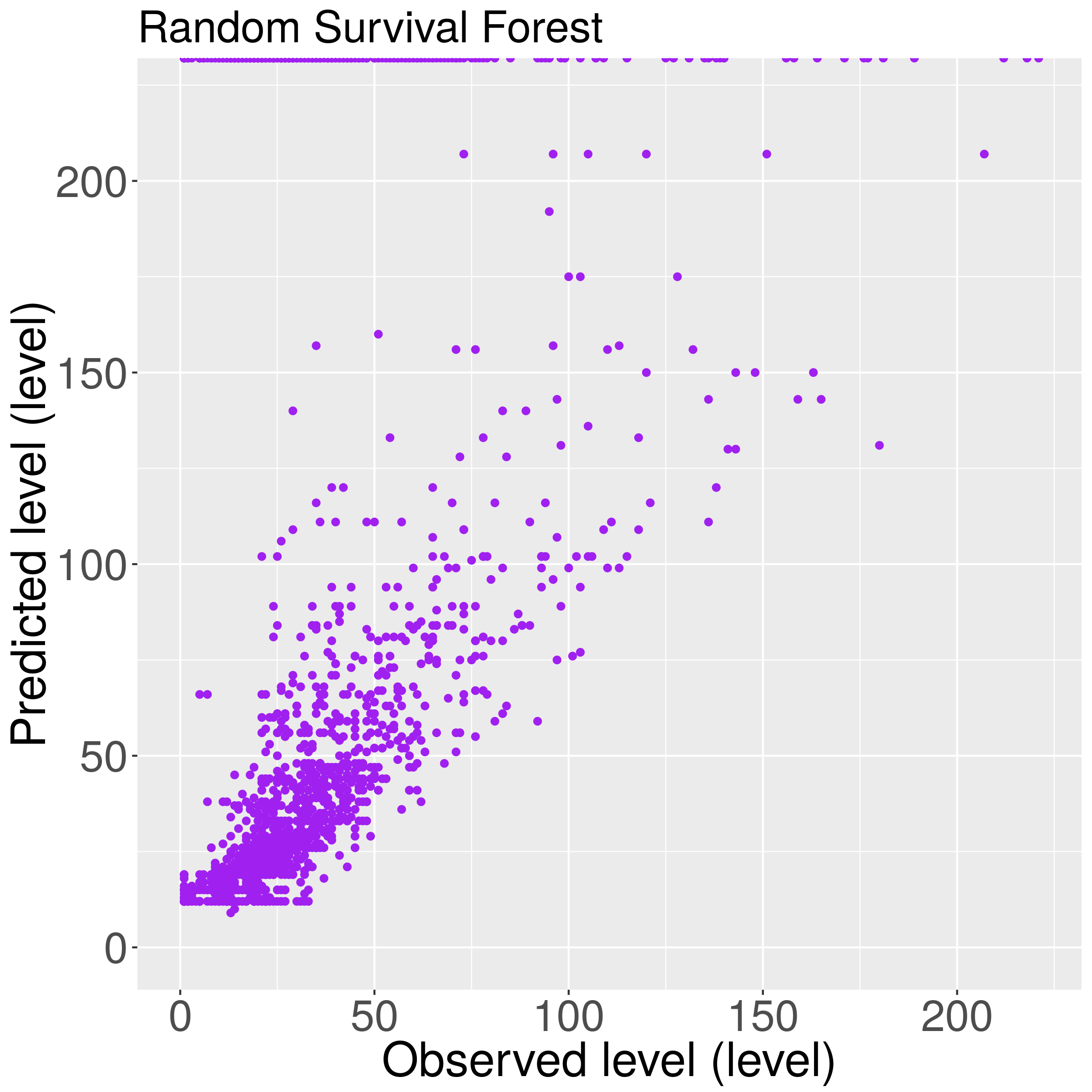} 
  \includegraphics[width=0.30\textwidth]{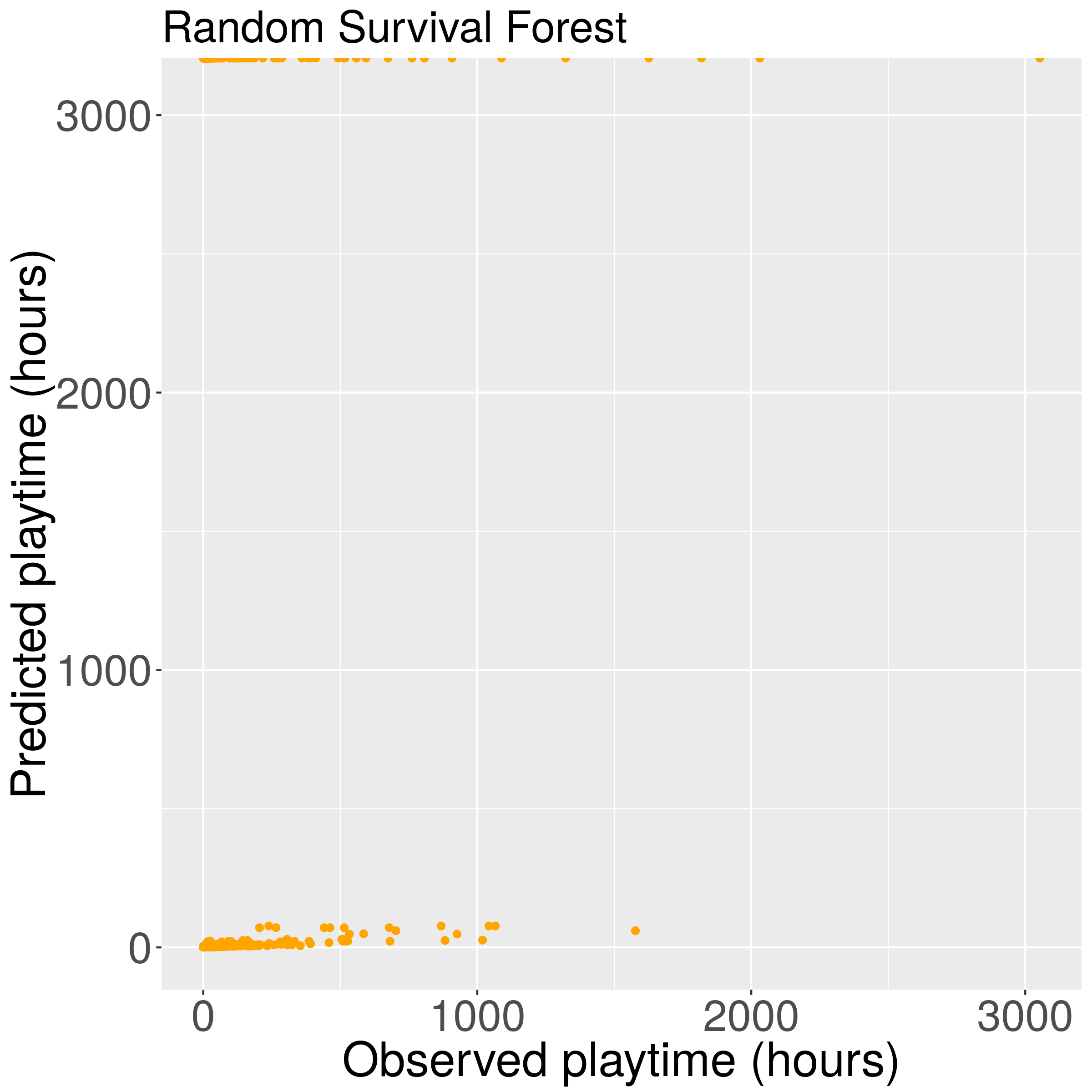}\\
  \includegraphics[width=0.30\textwidth]{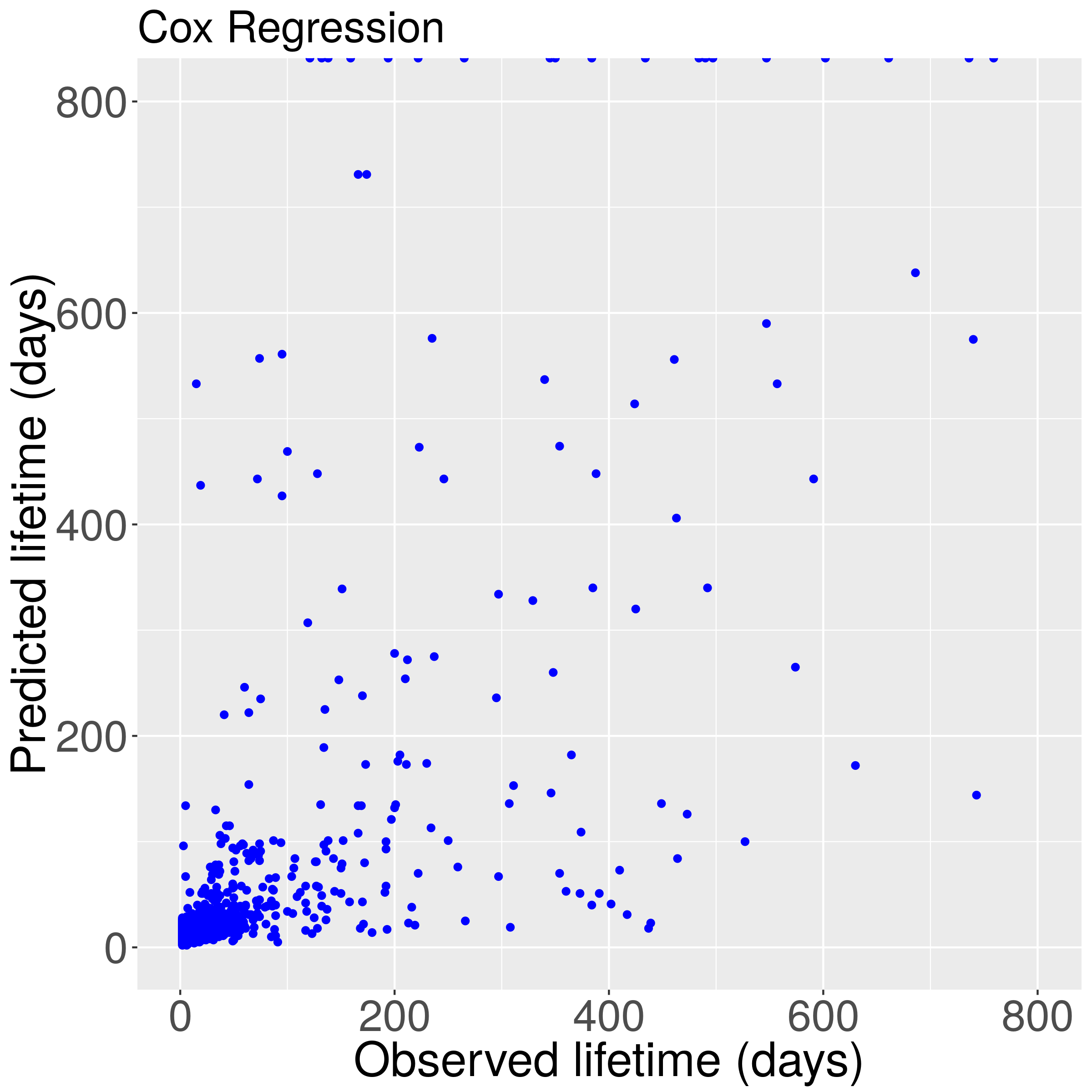} 
  \includegraphics[width=0.30\textwidth]{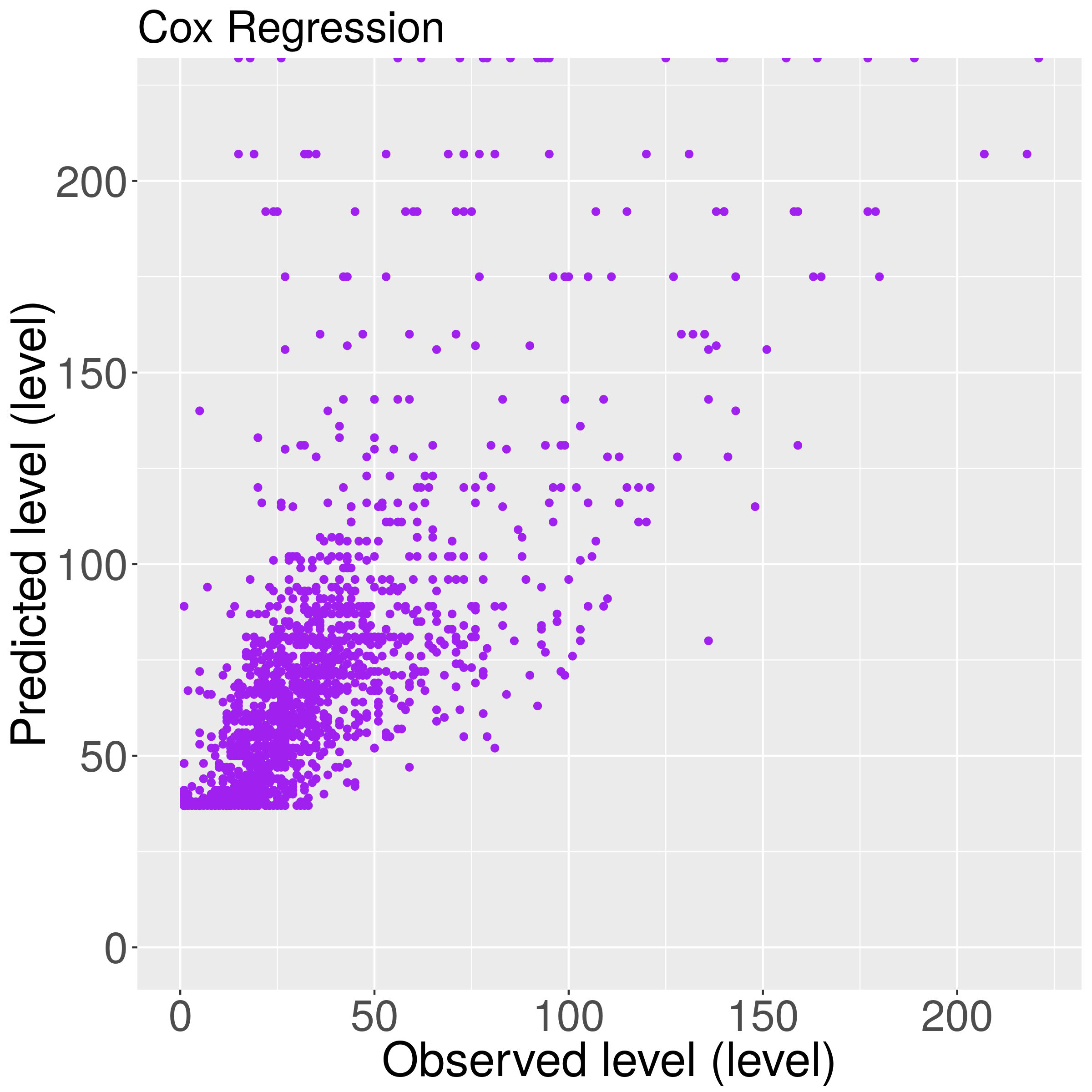} 
  \includegraphics[width=0.30\textwidth]{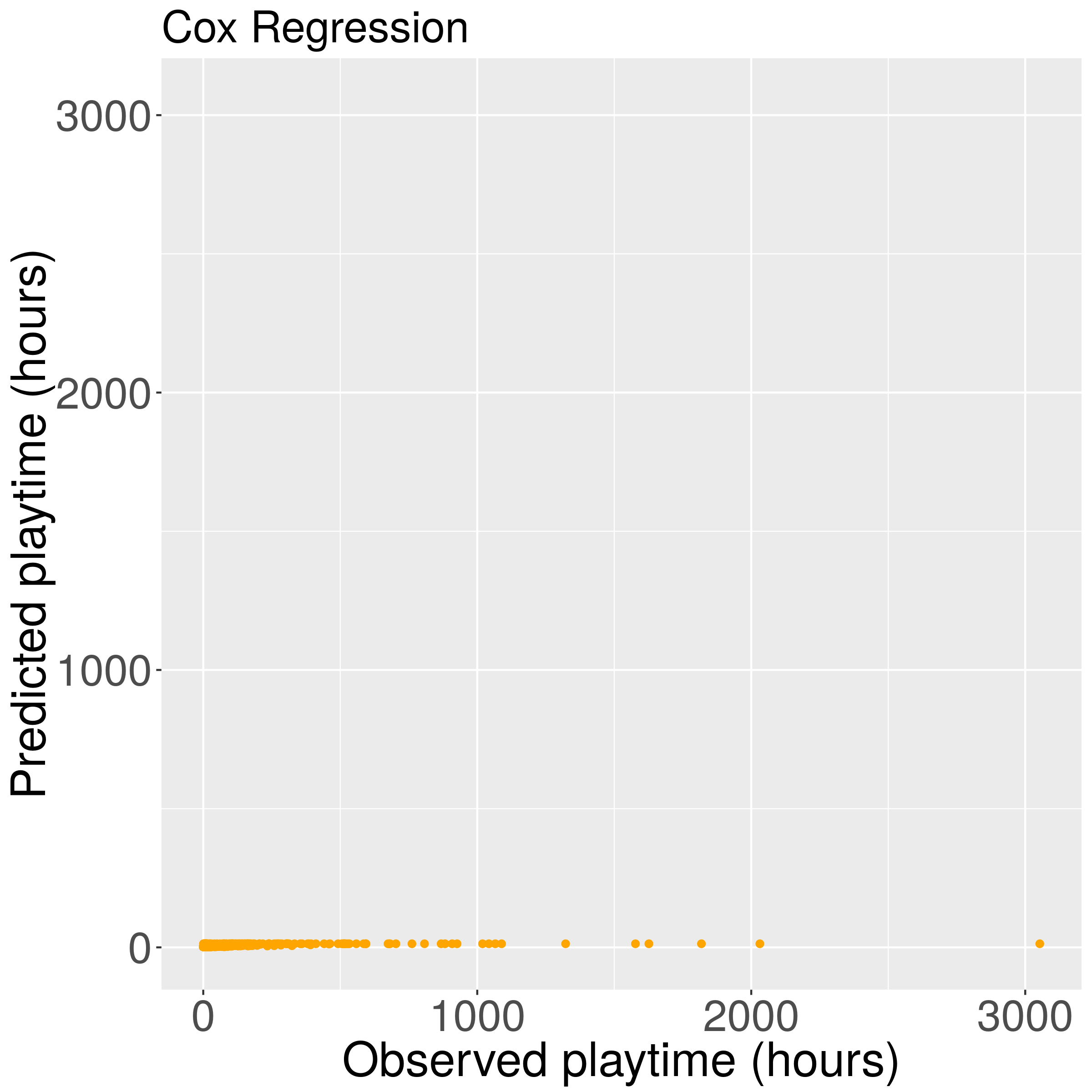}\\
\caption{Scatter plots of observed vs.\ predicted ``times'' for the occurrence of the event \emph{becoming a PU} for AoI players. We consider three different time measures---lifetime (left), game level (center) and playtime (right)---and three different models---conditional inference survival ensembles (top), random survival forest (middle) and Cox regression (bottom). Predictions correspond to the median survival values.}
\label{scatterPlots}
\end{figure*}

\begin{figure*}[ht!]
  \centering
  \includegraphics[width=0.30\textwidth]{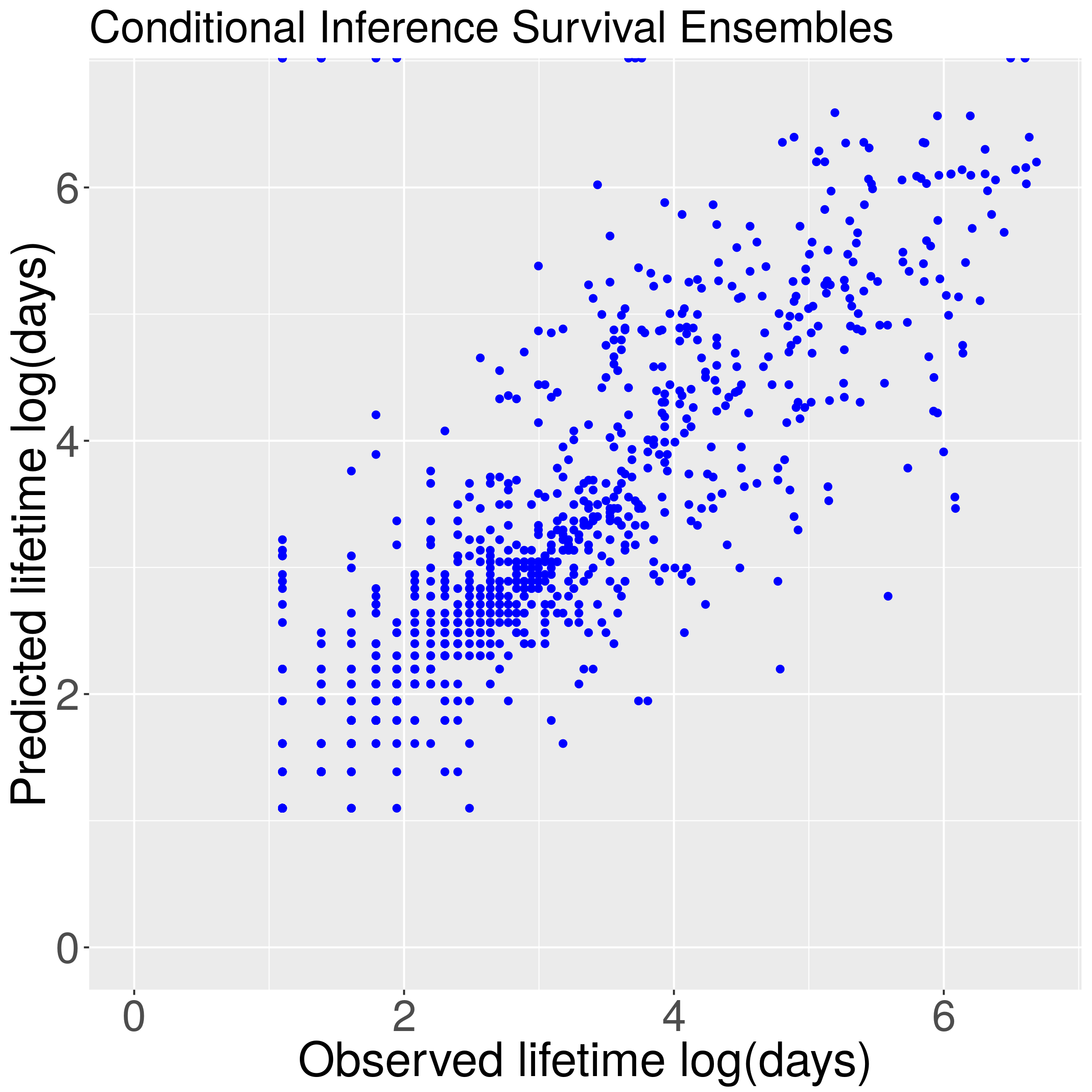} 
  \includegraphics[width=0.30\textwidth]{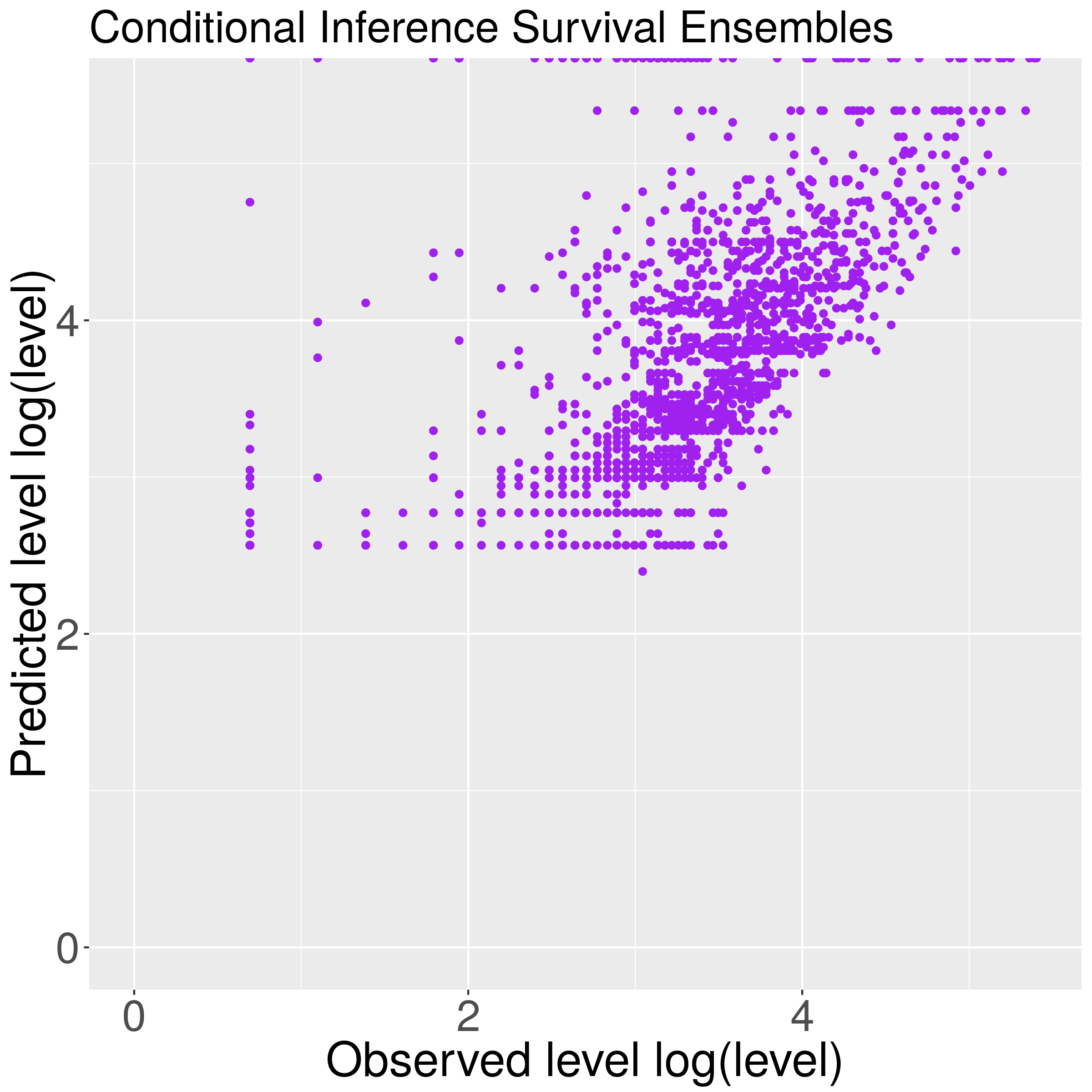} 
  \includegraphics[width=0.30\textwidth]{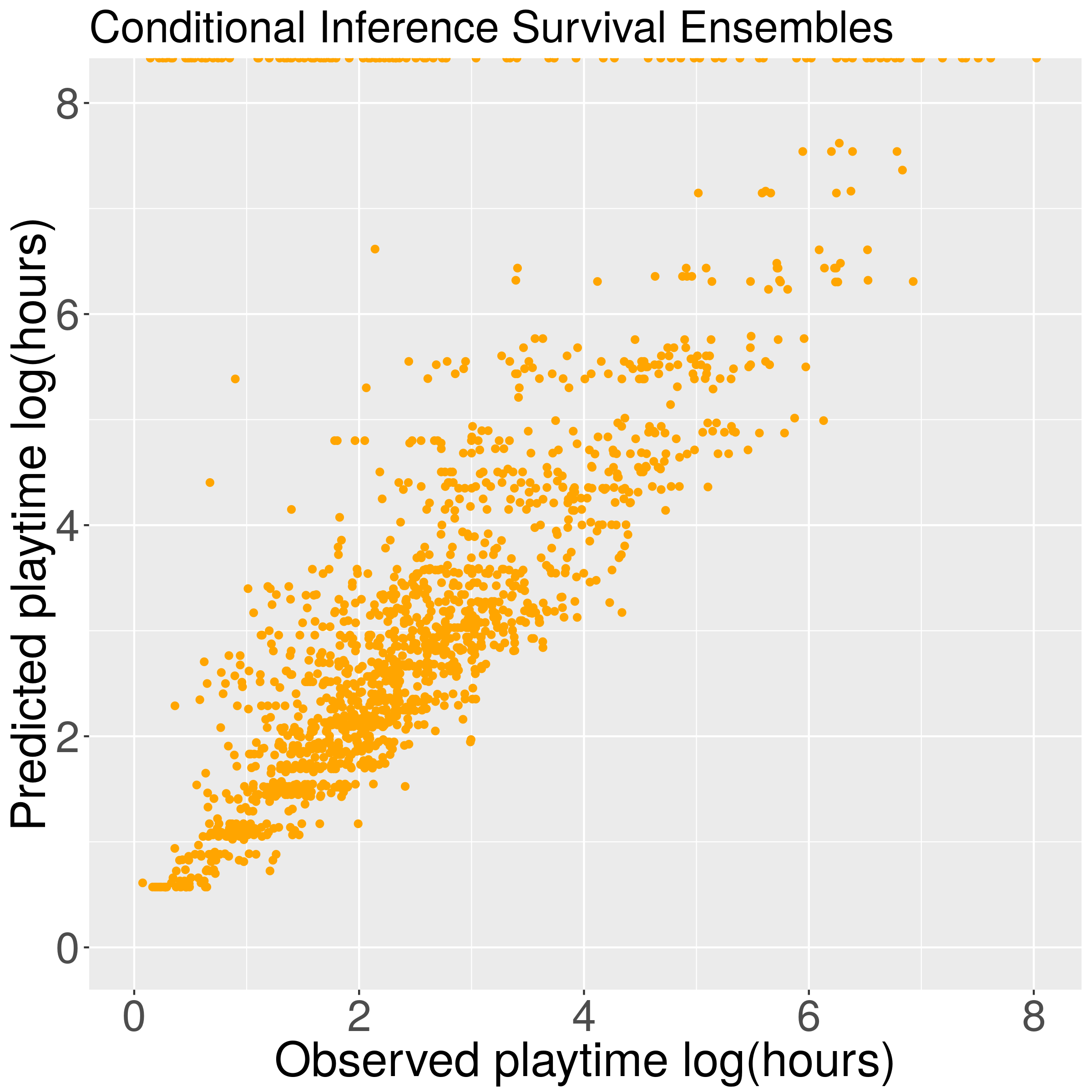}\\
  \includegraphics[width=0.30\textwidth]{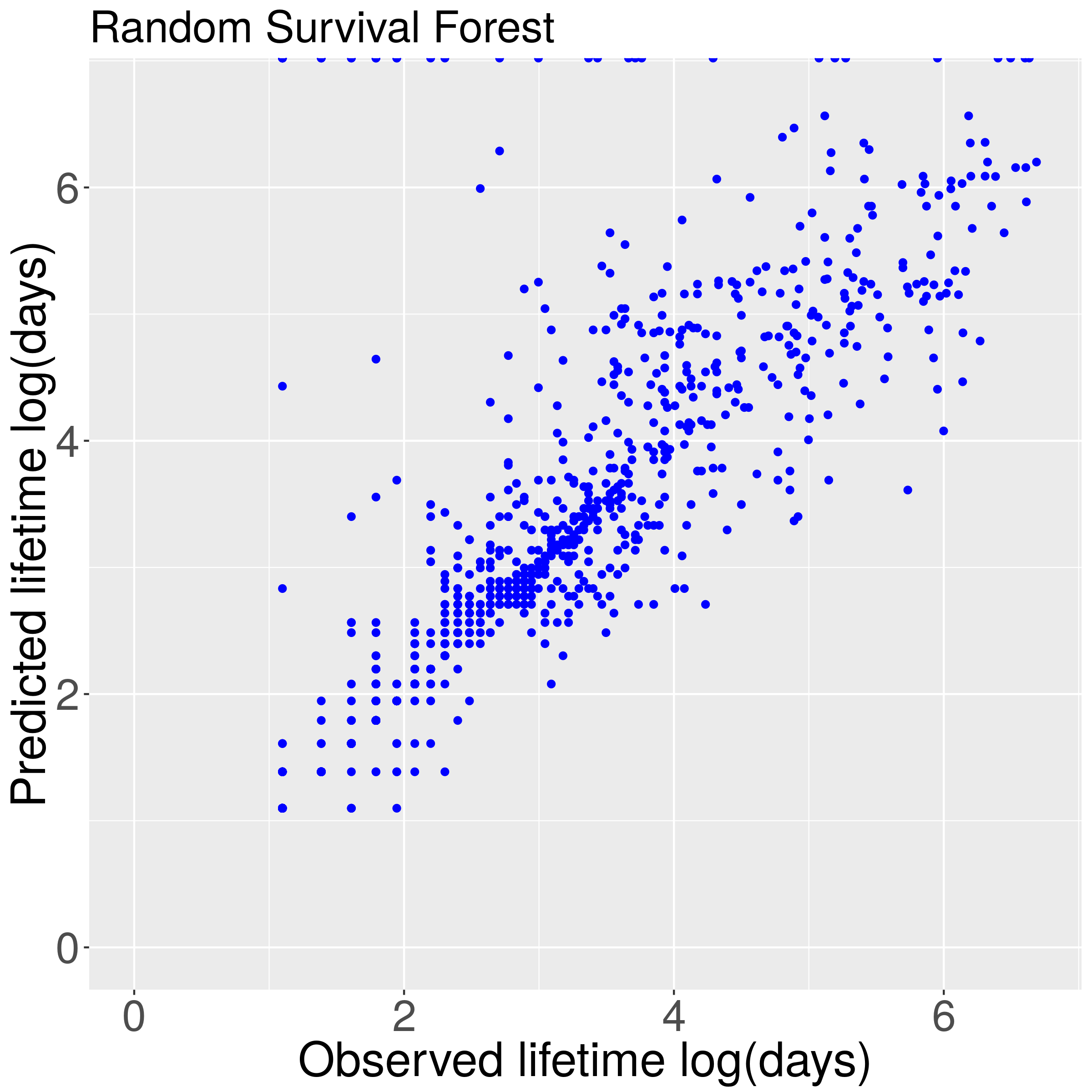} 
  \includegraphics[width=0.30\textwidth]{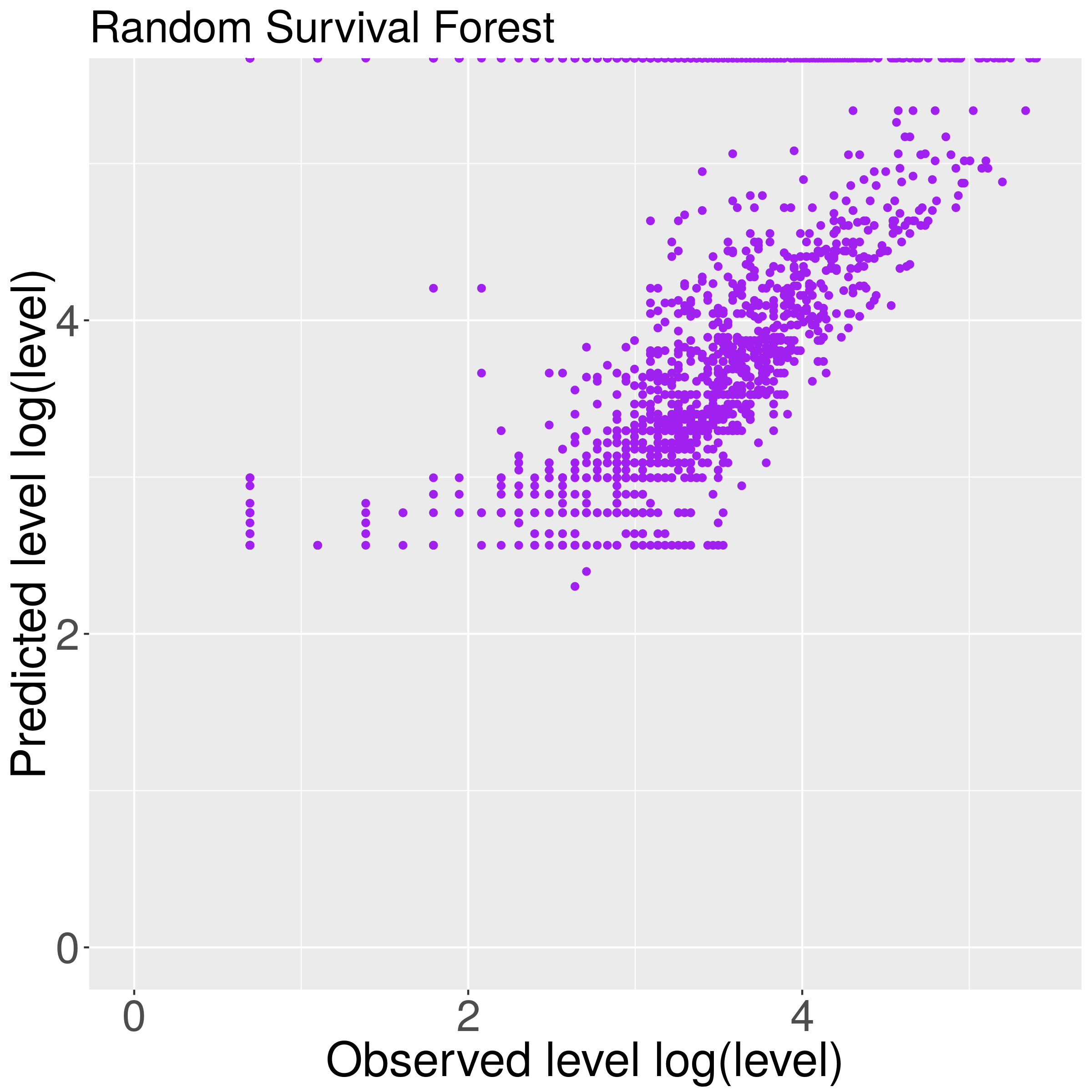} 
  \includegraphics[width=0.30\textwidth]{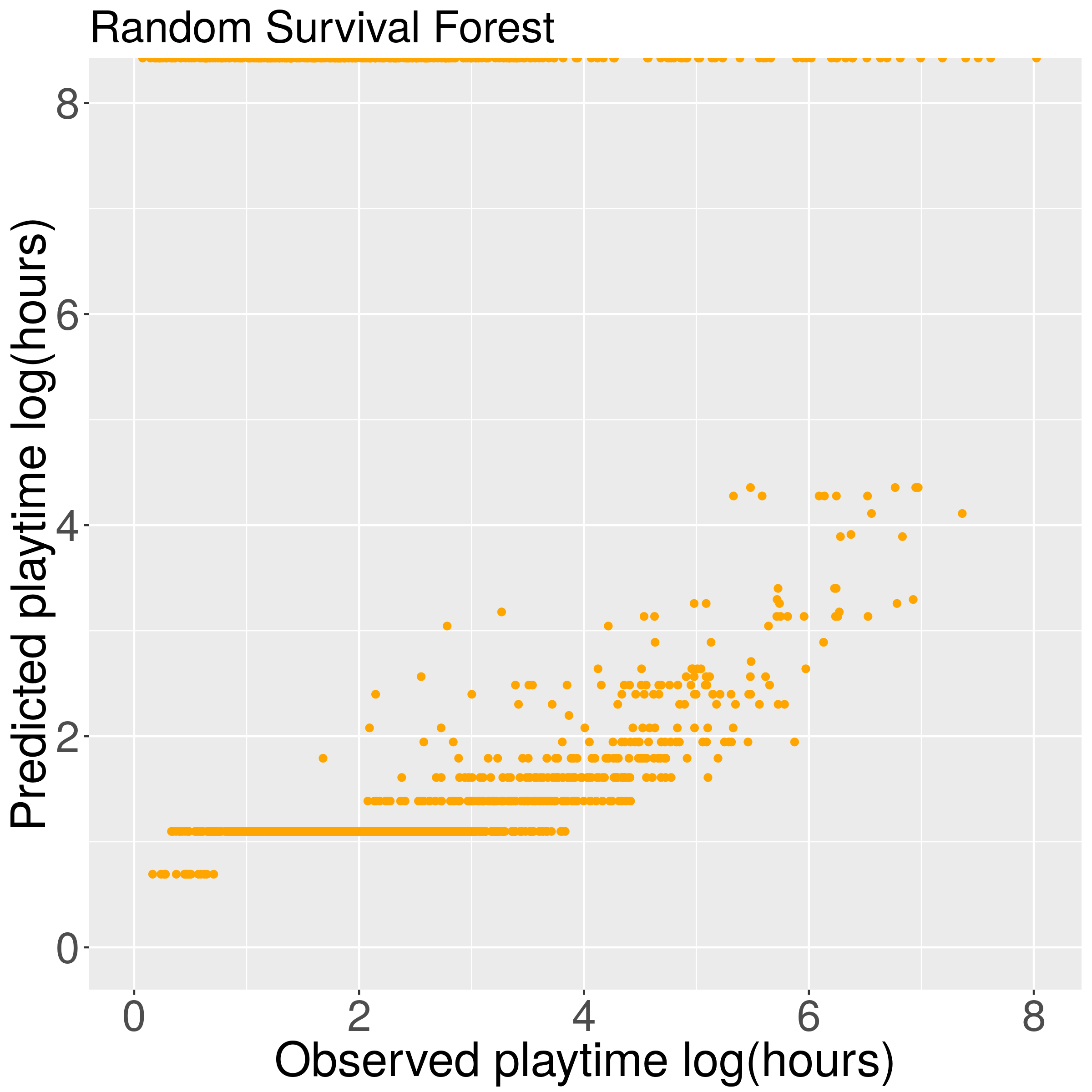}\\
  \includegraphics[width=0.30\textwidth]{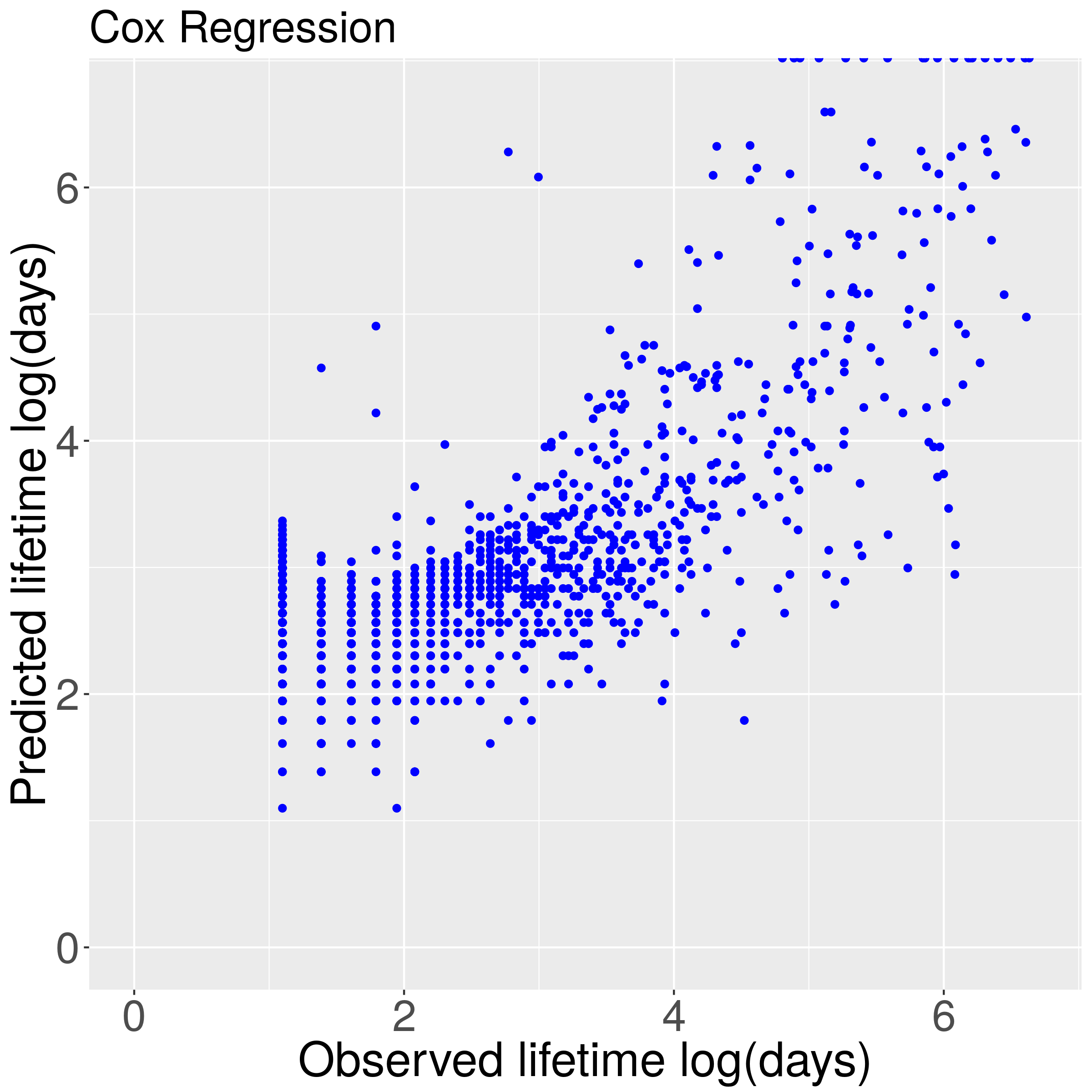} 
  \includegraphics[width=0.30\textwidth]{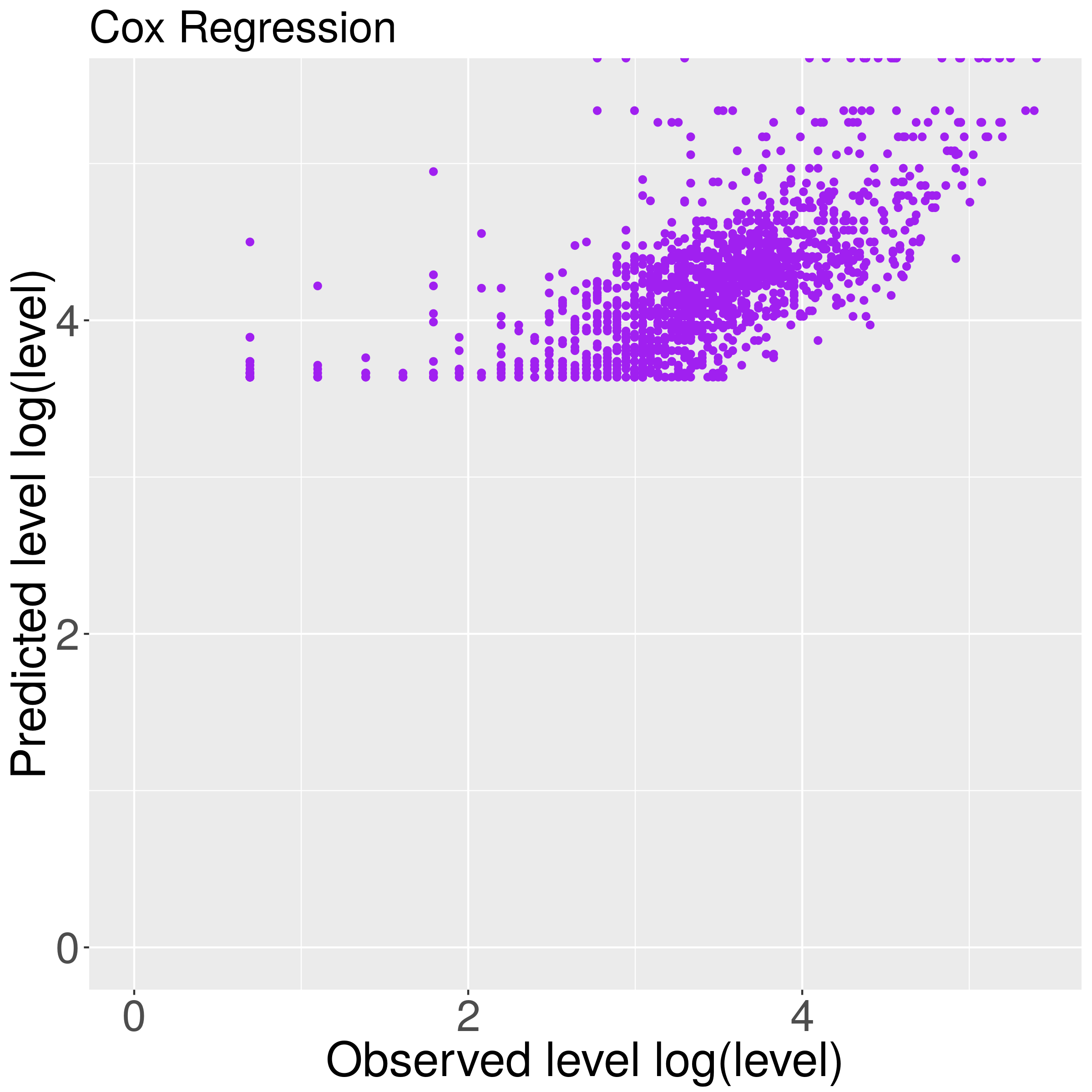} 
  \includegraphics[width=0.30\textwidth]{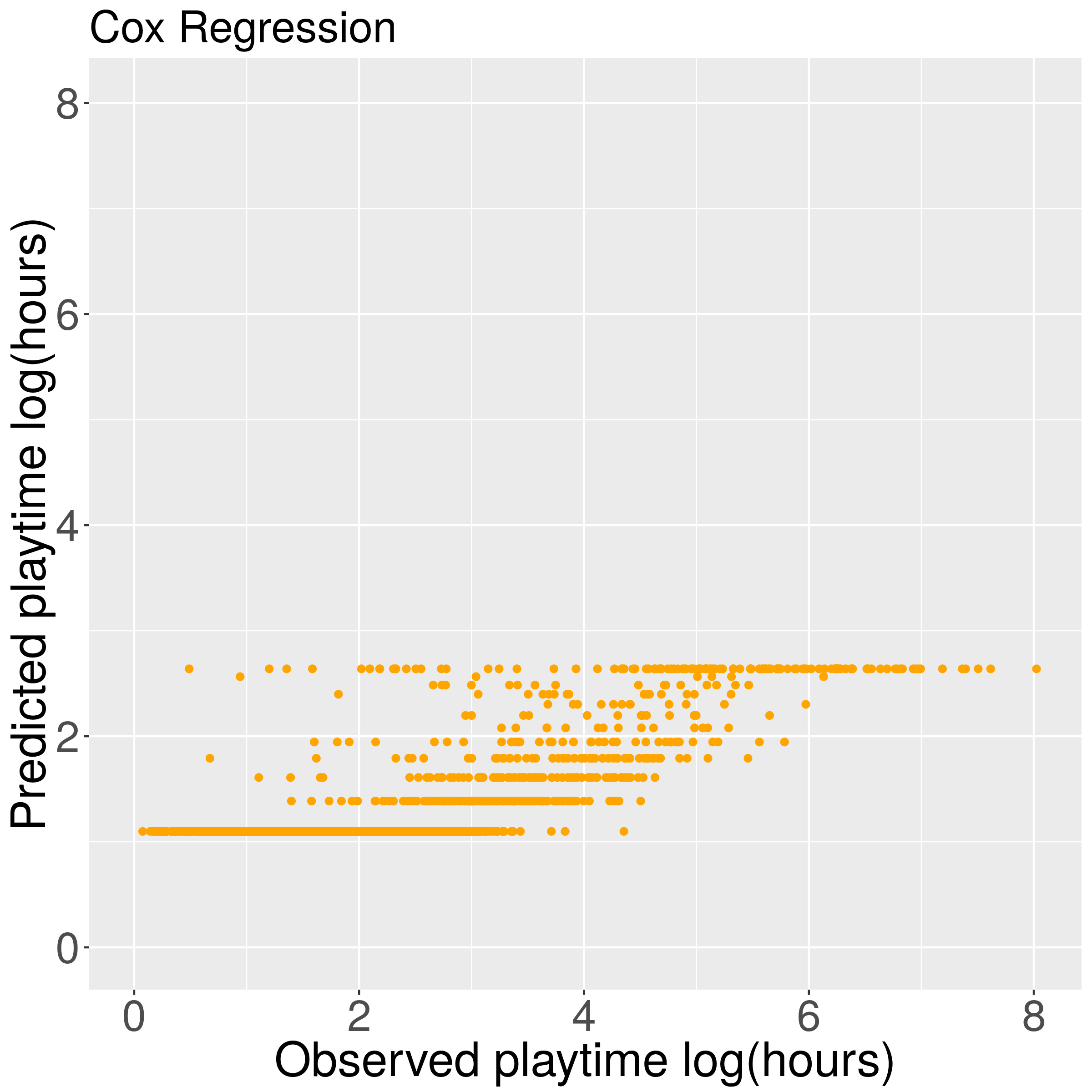}\\
\caption{Log-log scatter plots of observed vs.\ predicted ``times'' for the occurrence of the event \emph{becoming a PU} for AoI players. We consider three different time measures---lifetime (left), game level (center) and playtime (right)---and three different models---conditional inference survival ensembles (top), random survival forest (middle) and Cox regression (bottom). Predictions correspond to the median survival values. The logarithm transformation provides a close-up look at the spread of the data points (cf.~Figure~\ref{scatterPlots}).}
\label{scatterPlotslog}
\end{figure*}

The results for all different models and variables (lifetime, level and playtime) are summarized in Table~\ref{Errors}. Scatter plots comparing observed and predicted values for players that did become PUs are shown in Figure~\ref{scatterPlots}, whereas Figure \ref{scatterPlotslog} displays the corresponding log-log scatter plots. The latter are probably more illustrative, as using logarithms allows a close-up look at small values of the observed and predicted quantities while preventing a visual overpenalization by errors at large values. 

Considering the identification of potential PUs (regardless of when the conversion occurs) all models give accurate results, as inferred from the low rates of false negatives and false positives in Table~\ref{Errors}. All methods also provide reasonable predictions of when the conversion will take place in terms of the three variables, thus confirming the suitability of survival analysis to explore this problem. Overall results for the semi-parametric Cox regression model show relatively larger errors---across all variables and games---as compared to the ensemble approaches.

The three ensemble methods yield comparable results in general. It is worth noting that the model including competing risks does not outperform the others. This probably indicates that churn is not a competitive risk in nature, i.e. non-PUs with a high risk of churning very rarely become PUs and, conversely, players with a high probability of becoming PUs are normally not considering quitting the game. Taking churn into account does slightly reduce the rate of false positives, as would be expected, but produces a larger increase in the rate of false negatives (except for playtime in AoI). In regard to when conversions will occur (for those players that are indeed to become PUs) including competing risks results into less accurate predictions except for lifetime in GS. 

The RSF model yields slightly better lifetime and level predictions than conditional inference survival ensembles in both games, but performs significantly worse for playtime. In particular, conversions that occur after a very long playtime are only predicted by the conditional inference survival ensembles model, as can be seen in the scatter plots shown in Figures~\ref{scatterPlots} and \ref{scatterPlotslog}. This is of the utmost importance for the problem under consideration, as one of the obvious applications of this analysis would be to individually target potential PUs in order to accelerate their conversion. Even when the conversion happens after a short playtime, both the random survival forest and Cox regression models exhibit very obvious biases, yielding prediction values that are systematically lower than the actual outcomes. 

For level predictions, however, the RSF model produces better results across all scales in both games. The scatter plots in Figures~\ref{scatterPlots} and \ref{scatterPlotslog} also reveal the inability of all models to predict conversions in the first levels of the game---where player progression is typically very quick. This has however hardly any practical relevance: in these first stages of the game, conversions are almost immediate in terms of lifetime and playtime, so early detection of the potential of these players adds very little value. Similarly, although RSFs also provide overall better predictions for lifetime, this is due mainly to its better performance in cases when conversion takes place early on and which have thus limited impact for practical purposes. Note also that (although this effect is smaller in the case of the RSF method) all models are biased in that they tend to predict higher levels of conversion than actually observed. This is also the case for playtime predictions using conditional inference ensembles.

Scatter plots for GS are similar to those shown for AoI---as suggested by the results of Table~\ref{Errors}---and thus they are not included.

\begin{table*}
	\centering
	\caption{Validation results for all models and variables considered. (RMSLE: Root mean square logarithmic error.)}
	\setlength\tabcolsep{5pt}
	\begin{tabular}{cccccccccc} \toprule
    	\textbf{Age of Ishtaria (AoI)}                  &   \multicolumn{3}{c}{RMSLE} & \multicolumn{3}{c}{False Negatives} & \multicolumn{3}{c}{False Positives}\\ \cmidrule(r){2-4}\cmidrule(l){5-7}\cmidrule(l){8-10}
		Model                               & Lifetime & Level & Playtime &  Lifetime & Level & Playtime & Lifetime & Level & Playtime\\ \midrule
		Conditional inference survival ensembles & 0.54 & 0.69 & 0.47 & 0.27\%  & 0.84\% & 0.60\% & 3.68\%& 4.02\%& 4.02\% \\ 
		Random survival forest                   & 0.45 & 0.50 & 0.71 & 0.18\%  & 1.08\% & 1.01\% & 3.70\%& 3.32\%& 3.42\% \\ 		
		Random survival forest (competing risks) & 0.50 & 0.63 & 0.85 & 0.61\%  & 3.21\% & 0.58\% & 3.41\%& 1.17\%& 3.27\% \\ 
		Cox regression                           & 1.08 & 1.00 & 0.79 & 12.22\% & 1.69\% & 2.34\% & 3.75\%& 4.19\%& 2.30\% \\  \toprule
    	\textbf{Grand Sphere (GS)}                    &    \multicolumn{3}{c}{RMSLE} & \multicolumn{3}{c}{False Negatives} & \multicolumn{3}{c}{False Positives}\\ \cmidrule(r){2-4}\cmidrule(l){5-7}\cmidrule(l){8-10}
		Model                               & Lifetime & Level & Playtime & Lifetime & Level & Playtime  & Lifetime & Level & Playtime\\ \midrule
		Conditional inference survival ensembles & 1.00 & 0.77 & 0.48 & 1.74\% & 0.58\% & 1.31\% & 1.54\%& 3.09\%& 2.97\% \\ 
		Random survival forest                   & 0.58 & 0.63 & 0.79 & 1.78\% & 1.07\% & 1.17\% & 1.62\%& 2.47\%& 2.38\% \\ 
		Random survival forest (competing risks) & 0.34 & 0.92 & 0.83 & 2.42\% & 2.89\% & 3.39\% & 1.07\%& 0.59\%& 2.30\% \\ 
		Cox regression                           & 2.71 & 1.23 & 0.85 & 3.07\% & 3.66\% & 3.46\% & 1.70\%& 3.36\%& 3.11\% \\ \bottomrule
	\end{tabular}
	\label{Errors}
\end{table*}

\section{Summary and Conclusion}
\label{conclusion}
Our results show that survival analysis is a suitable framework to study user conversion in video games. We implemented several survival analysis methods, including three ensemble-based approaches, to determine the time, number of levels and accumulated playtime that non-paying players need to become PUs in two different free-to-play games. Historical data is included in the models at the individual level, as the aim of this work is to provide prediction results for each user. 

All models are very good at detecting potential PUs and provide fairly accurate time-to-event predictions in terms of days after first login, game level and playtime. Ensemble models outperform the classical semi-parametric Cox regression model across most validation metrics, variables and games. They are also particularly well suited for operational settings, as they can be easily parallelized and thus admit a scalable implementation.

Among the different ensemble approaches considered, the RSF method yields slightly better predictions in terms of lifetime and level, but critically fails at predicting playtime for those players who only start purchasing after having played for a very long time. Including churn as a competing risk does not have any clear positive impact. Moreover, RSFs are notorious for their proneness to introducing biases, as they favour variables with many splitting points. These results point to conditional inference survival ensembles as the most viable model in controlled production settings.

This work represents a step toward the personalization of the game experience in that it serves to target players individually, not only based on their current or past actions but also on their expected future behavior. Game developers and planners could use these methods to automatically determine who is likely to become a premium player and when she is likely to start behaving as such. This information can be then used to tailor the game experience of players with several goals in mind. Actions can be taken on players that have potential to become PUs to ensure they remain long enough in the game for the conversion to take place. Actions can be also taken to motivate each user at the precise moment or adequate stage of the game instead of targeting them too early on, when, for example, notifications or discounts are more likely to bother and disengage the players than to produce the conversion. These predictions also bring attention to those players who are not expected to become PUs in the near future, so as to try to accelerate their conversion if and when possible.

Future extensions of this work include applying the same approach to identify potential top spenders among the already existing PUs, and to detect conversions between different types of purchasing behavior, which should enable further personalization and increased monetization. For example, while for frequent spenders with low average outlay the goal would be to increase the latter, for players that seldom make purchases, efforts directed toward raising their purchasing frequency will probably be more effective.

\section{Software}
\label{software}

All analyses were performed using R version 3.4.4 for Linux and the following packages from the Comprehensive~R Archive Network (CRAN): {\em party} 
(version 1.3-0) \cite{hothorn2010party,hothorn2015package}, {\em survival} (version 2.42-6) \cite{therneau2015package}, {\em survminer} (version 0.4.3) \cite{kassambara2017survminer,kassambara2017package}, {\em ROCR} (version 1.0-7) \cite{sing2005rocr,sing2007package}, {\em randomForestSRC} (version 2.8.0) \cite{ishwaran2019package} and {\em peperr} (version 1.1-7) \cite{porzelius2019package}. 

\begin{acks}
\label{acknowledgements}
We thank Javier Grande for his careful review of the manuscript.
\end{acks}

\bibliographystyle{ACM-Reference-Format}
\bibliography{main}

\end{document}